\documentclass[accepted]{uai2021} 

\usepackage{natbib} 
\usepackage{microtype}
\usepackage{graphicx}
\usepackage{subfigure}
\usepackage{booktabs} 

\usepackage{hyperref}

\usepackage{url}            
\usepackage{amsfonts}       
\usepackage{nicefrac}       

\usepackage{amsmath}
\usepackage{amsthm}

\def\eqref#1{Equation~\ref{#1}}			
\def\figref#1{Figure~\ref{#1}}			
\def\tabref#1{Table~\ref{#1}}			
\def\secref#1{Section~\ref{#1}}			
\def\asmref#1{Assumption~\ref{#1}}		
\def\thmref#1{Theorem~\ref{#1}}		

\newtheorem{dff}{Definition}
\newtheorem{thm}{Theorem}

\newtheorem{asm}{Assumption}

\usepackage{cleveref}
\crefrangeformat{figure}{Figure #3#1#4--#5#2#6}

\makeatletter
\newcommand*{\indep}{%
  \mathbin{%
    \mathpalette{\@indep}{}%
  }%
}
\newcommand*{\nindep}{%
  \mathbin{
    \mathpalette{\@indep}{\not}
  }%
}
\newcommand*{\@indep}[2]{%
  \sbox0{$#1\perp\m@th$}
  \sbox2{$#1=$}
  \sbox4{$#1\vcenter{}$}
  \rlap{\copy0}
  \dimen@=\dimexpr\ht2-\ht4-.2pt\relax
  \kern\dimen@
  {#2}%
  \kern\dimen@
  \copy0 
} 
\makeatother

\usepackage{mwe}

\title{Improving Efficiency and Accuracy of Causal Discovery\\ Using a Hierarchical Wrapper}

\author[ ]{Shami Nisimov}
\author[ ]{Yaniv Gurwicz}
\author[ ]{Raanan Y.~~Rohekar}
\author[ ]{Gal Novik}

\affil[ ]{Intel Labs}

\begin{document}
\maketitle

\vskip 0.3in




\begin{abstract}
Causal discovery from observational data is an important tool in many branches of science. Under certain assumptions it allows scientists to explain phenomena, predict, and make decisions. In the large sample limit, sound and complete causal discovery algorithms have been previously introduced, where a directed acyclic graph (DAG), or its equivalence class, representing causal relations is searched. However, in real-world cases, only finite training data is available, which limits the power of statistical tests used by these algorithms, leading to errors in the inferred causal model. This is commonly addressed by devising a strategy for using as few as possible statistical tests.
In this paper, we introduce such a strategy in the form of a recursive wrapper for existing constraint-based causal discovery algorithms, which preserves soundness and completeness. It recursively clusters the observed variables using the normalized min-cut criterion from the outset, and uses a baseline causal discovery algorithm during backtracking for learning local sub-graphs. It then combines them and ensures completeness. By an ablation study, using synthetic data, and by common real-world benchmarks, we demonstrate that our approach requires significantly fewer statistical tests, learns more accurate graphs, and requires shorter run-times than the baseline algorithm.
\end{abstract}

\section{Introduction}
\citep{glymour2019review}
A fundamental task in various disciplines of science is to discover causal relations among domain variables \citep{glymour2019review, shen2020challenges}. In many cases, the causal relations can be properly represented by a DAG \citep{pearl2009causality}. Then, by interpreting this causal DAG as a statistical model, many of these causal relations can be discovered using observational data alone \citep{spirtes2000, pearl1991theory, peters2017elements}, known as causal discovery. 
In constraint-based causal discovery algorithms, statistical independence is tested between pairs of variables conditioned on subsets of the remaining domain variables \citep{spirtes2000, colombo2012learning, claassen2013learning, tsamardinos2006max, yehezkel2009rai, cheng2002learning}. As not all causal relations can be discovered purely from these statistical tests using observational data, these algorithms return an equivalence class of the true underlying DAG. Nevertheless, constraint-based algorithms are generally proven to be asymptotically correct. In this paper, we will consider this family of algorithms.

In most real-world cases, limited observational data is available and statistical tests are prone to errors. Moreover, statistical tests for conditional independence (CI) often suffer from the curse-of-dimensionality. Tests with large condition sets are more prone to errors than tests with smaller condition sets. Thus, a common principle in constraint-based algorithms is to derive the next CI tests to perform, from the result of previous CI tests of smaller condition sets \citep{spirtes2000}.
Another challenge is that learning causal DAGs from observed data is NP-hard \citep{chickering2004large}. The number of possible DAGs grows super-exponentially with the number of domain variables, posing a serious limitation on the expected computational complexity of algorithms for real-world applications. On one hand, it is assumed that enough data points are available such that the statistical test results will be reliable, but on the other hand, the computational complexity of these statistical tests increases with the number of data points. Thus, the number of statistical tests commonly serves as a measure of computational complexity.

The common approach to address these problems is to reduce the overall number of CI tests required by the algorithm, and favor those that have greater statistical power. In this paper, we propose a wrapper---hierarchical clustering for causal discovery (HCCD)---for existing casual discovery algorithms, referred in the paper as baseline algorithms. This wrapper recursively clusters the domain variables, thus limiting the condition set size of CI tests within each cluster, which alleviates the curse-of-dimensionality. That is, the wrapper relies on the level of correlation between variables, as apposed to the Boolean result of the statistical CI tests. 
Using spectral clustering, sub-domains are derived with respect to the \emph{relative} correlation between variables, recursively. Once a (sub-) domain cannot be divided into smaller sub-domains, a baseline causal discovery algorithm is called. Tracing back from the recursion, the causal graphs for each sub-domain are merged and the baseline algorithm is called again on the merged graph for the edges between sub-domains (inter-domain), retaining edges within each sub-domain (intra-domain). The proposed wrapper improves accuracy in common finite datasets while preserving the soundness and completeness of the baseline algorithm.

\begin{figure*}
    \centering
    \subfigure[]{\includegraphics[width=0.45\textwidth]{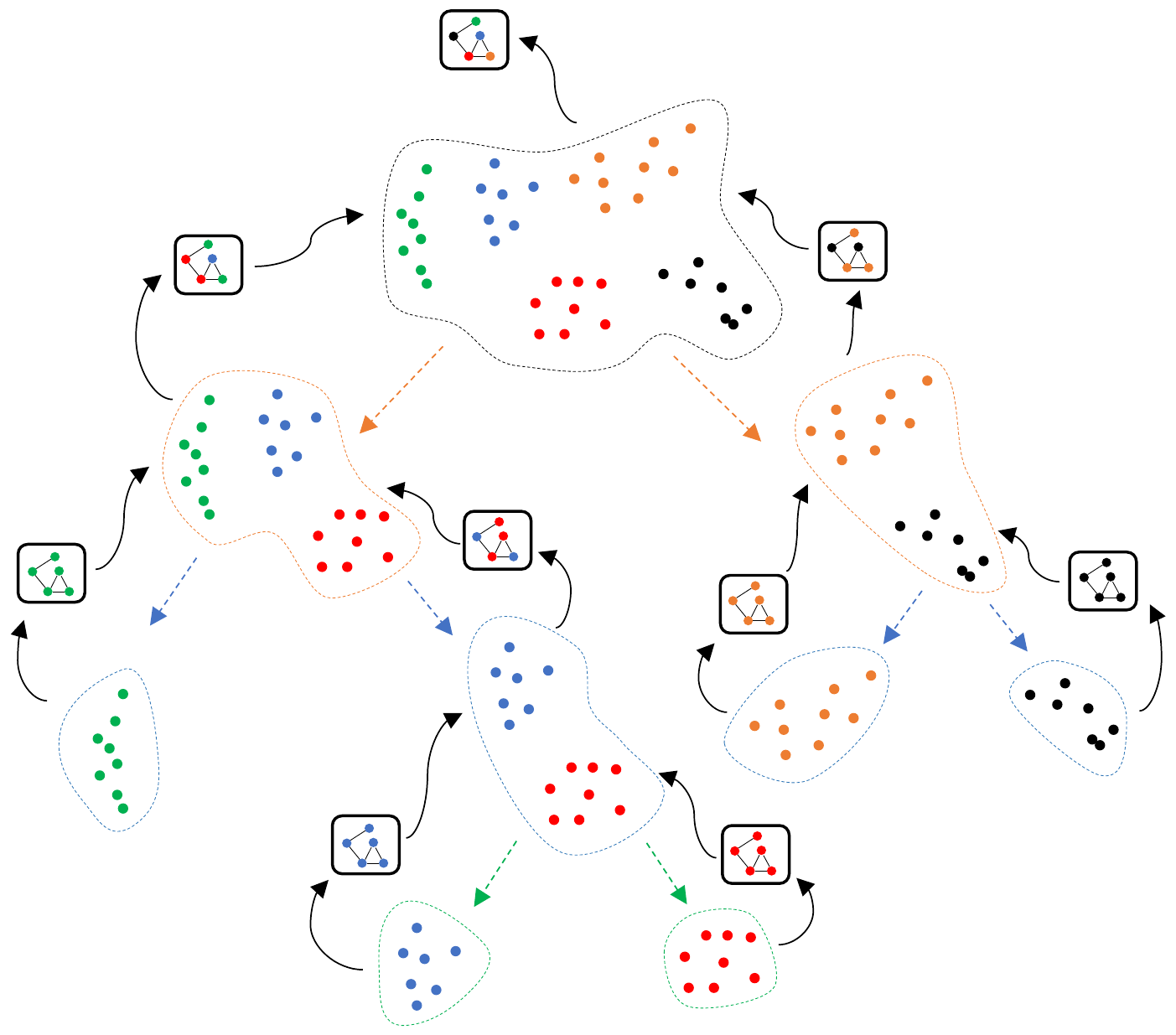}} 
    \subfigure[]{\includegraphics[width=0.53\textwidth]{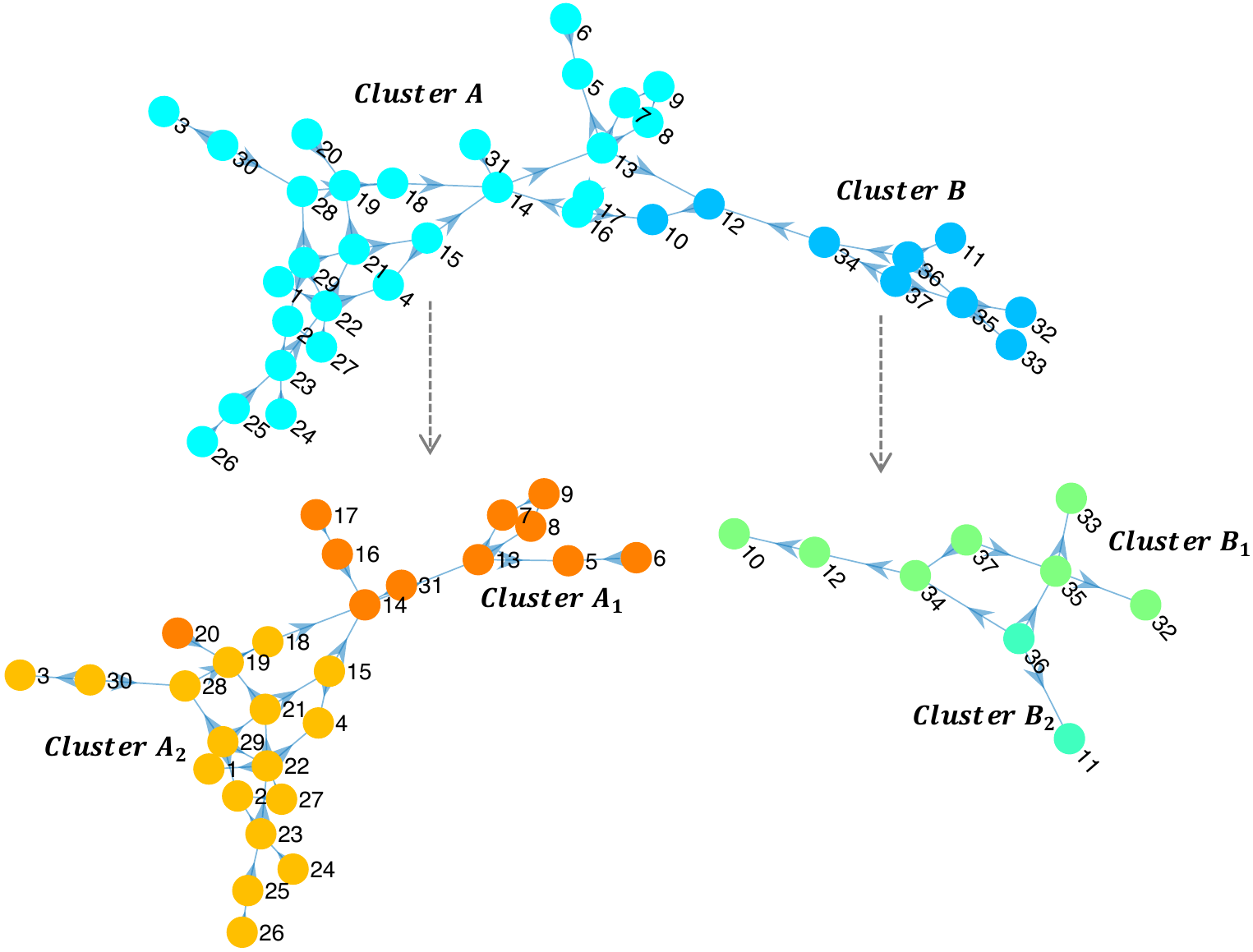}}
    \vskip -0.02in
    \caption{(a) An illustration of a top-down 2-way clustering of feature set followed by and a bottom-up causal discovery. The domain variables are clustered hierarchically. Then, from the leaves upwards, causal discovery (shown as a colored graph) is applied disjointly to the variables in each cluster, and then the resultant graphs are unified at their parent cluster, conditioned on their white-lists as depicted by those graphs. This process is backtracked until the root of the cluster tree. Best viewed in color. (b) An example to the a top-down 2-way clustering of the ALARM dataset's domain variables. In the first level, HCCD creates 2 clusters $\{A,B\}$. Then, for each cluster a recursive call is evaluated, and clusters $\{A_1, A_2\}$ and $\{B_1, B_2\}$ are created, respectively. The nodes within each cluster are highlighted by a different color, and presented on the ground truth structure. Best viewed in color.}
    \label{fig:method_illustration}
    \vskip -0.08in

\end{figure*}

\section{Background}

Constraint-based algorithms for causal discovery rely on the correctness of statistical tests for conditional independence. In practice, as only limited data is available, these tests are prone to errors, and often suffer from the curse-of dimensionality. The PC algorithm \citep{spirtes2000} iteratively refines an initial fully-connected graph. In each iteration, connected nodes are tested for independence conditioned on a subset of their neighbors, where this subset is restricted to a constant size. The edge is removed if an independence is found. The restriction on the condition set size is increased by one in the next iteration. Thus, this approach has an advantage when CI tests with smaller condition sets are more reliable than CI tests with larger condition sets.
The RAI algorithm \citep{yehezkel2009rai} follows this approach but relies heavily on information from CI tests with smaller condition sets. It orients the graph's edges and decomposes it into sub-graphs before additional CI testing. Thus, errors in earlier stages may cause errors in later stages.  
Other works \citep{cai2017sada, aliferis2010local, xie2008recursive} also leverage divide-and-conquer or local-search strategies in a hierarchical or recursive way, and report improved results by partitioning the nodes into subsets and learning local structure for each.

A line of work \citep{sondhi2019reduced, chickering2015selective} propose to leverages properties of the graph to improve running time. Previously, it was shown that relying on correlation level between pairs of variables, in addition to the Boolean result of the CI tests, can reduce the overall number of CI tests and improve accuracy. The TPDA algorithm \citep{cheng2002learning}, having a complexity of $O(n^4)$ ($n$ is the number of nodes), relies on the monotone-DAG-faithfulness assumption. It assumes that the mutual information between any pair of nodes cannot decrease by opening more dependency inducing paths. Nevertheless, although TPDA has lower complexity than algorithms that do not utilize the correlation level among variables, it performs more CI tests having large condition sets, rendering it unstable for limited data \citep{tsamardinos2006max}.
Recently, it was proposed to utilize inhomogeneity in the domain as a heuristic for improving accuracy and speed of existing causal discovery algorithms \citep{pashami2018causal, zhang2018learning}. For example, TSCB \citep{zhang2018learning} is a 2-step wrapper algorithm that first clusters the domain variables invoking an existing causal discovery algorithm for each cluster, and then applies the same causal discovery algorithm to inter-cluster edges. However, it is not clear under which conditions clustering-based wrappers retain soundness and completeness of the baseline algorithm, and under which conditions they are faster and learn a more accurate graph than the baseline algorithm. 

In this paper we discuss the implication of domain variables clustering on the soundness and completeness of causal discovery. We also discuss the properties of such clustering that may reduce or increase the probability of errors and the overall efficiency.

\section{Variables Clustering for Causal Discovery}

As discussed, dividing the set of variables into smaller subsets can be appealing for causal discovery algorithms. However, in the common scenario where the underlying causal DAG is connected, an optimal clustering, from which a causal discovery algorithm can benefit, is not clear.

Constraint-based algorithms often rely on the causal Markov and faithfulness assumptions \citep{spirtes2000}. A probability distribution $P$ and a DAG $\mathcal{G}$ are said to be faithful to one another if in $P$, variables $A$ and $B$ are independent conditioned on set $\boldsymbol{Z}$ if and only if $A$ and $B$ are d-separated by $\boldsymbol{Z}$ in $\mathcal{G}$, $A\indep B | \boldsymbol{Z}$. It is then key in constraint-based algorithms to identify conditional independence relations for constructing the underlying graph. Let $\mathrm{Alg}$ be a causal discovery algorithm. Let $\mathrm{ClustCD}$ (Cluster Causal Discovery) be the following procedure. 
1) Given observed data for domain variables $\boldsymbol{X}=\{A, B, \ldots\}$, partition $\boldsymbol{X}$ into $k$ disjoint subsets $\boldsymbol{X}_1, \ldots, \boldsymbol{X}_k$, i.e., $\cup_{i=1}^{k}\boldsymbol{X}_i=\boldsymbol{X}$ and $\boldsymbol{X}_i\cap\boldsymbol{X}_j=\emptyset, \forall i\neq j$.
2) Call $\mathrm{Alg}$ for each $\boldsymbol{X}_i$ (intra-cluster).
3) Call $\mathrm{Alg}$ for edges between any pair $(A,B)$ such that $A\in\boldsymbol{X}_i$ and $B\in\boldsymbol{X}_j$, for all $i,j\in\{1,\ldots,k\}, i\neq j$ (inter-cluster).

\begin{thm}\label{thm:twostep}
If $\mathrm{Alg}$ is a sound and complete causal discovery algorithm, then procedure $\mathrm{ClustCD}$ is sound for $\boldsymbol{X}$, but not complete.
\end{thm}
\begin{proof}
\vskip -0.15in
The proof is given in Appendix A.
\vskip -0.3in
\end{proof}

Given a probability distribution $P$ faithful to DAG $\mathcal{G}$, a complete algorithm can identify from observed data of $\boldsymbol{X}_i$ the conditional independence relation between a pair $A,B\in\boldsymbol{X}_i$, not adjacent in $\mathcal{G}$, if there is at least one separating set, $Z\in\boldsymbol{X}_i$, i.e. $A\indep B | Z$, where $A, B, Z$ are disjoint sets.
In general, it is not guaranteed that a partition of $\boldsymbol{X}$ into disjoint subsets (clustering) exists such that at least one separating set for every pair of conditionally independent variables are in the same cluster. Of course, there are cases where such a clustering does exists; for example, the clustering $\{A\}, \{B\}, \{C,D,E\}$ when the underlying graph is $A\rightarrow D \leftarrow C \rightarrow E \leftarrow B$.
Now, consider the case of two clusters. In one extreme, the first cluster contains a single variable, and the second cluster contains the remaining variables. In such a case, the expected number of undetectable intra-cluster independence relations is minimized. However, the complexity of the number of independence tests is maximal. 
On the other extreme, the two clusters have equal number of variables. This minimizes the complexity of the number of independence tests performed by the algorithm. For example, the complexity of the PC algorithm is $O(n^m)$, ($m$ is the maximal in-degree), so if one cluster has $n_1$ variables and the other $n-n_1$, then $O(n_1^m) + O((n-n_1)^m)$ is minimal for $n_1=\nicefrac{n}{2}$. However, the expected number of undetectable intra-domain independence relations is maximal.
A clustering method used by procedure $\mathrm{ClustCD}$ should balance minimizing the number of undetectable independence relations and the complexity of CI tests. For reducing the number of CI tests in the typical case, we assume that unconnected pairs of variables in $\mathcal{G}$ are more correlated to nodes of the minimal separating set, relative  to other nodes. 
\begin{asm}
Let $I$ be a pairwise symmetric correlation function. 
For every disjoint pairs of nodes $(X,Y)$ in the true underlying graph, such that $X \indep Y | \boldsymbol{Z}$, where $\boldsymbol{Z}$ is a minimal separating set, there exists $\boldsymbol{V} \subset \boldsymbol{X} \setminus (\{X,Y\}\cup\boldsymbol{Z})$, called a redundant set, such that 

\vskip -0.3in
\begin{equation*}
\begin{split}
\min_{Z\in\boldsymbol{Z}}\left[\max\left[I(X,Z), I(Y,Z)\right]\right] \;\ge\; I(X,Y) \;>\; \\ \min_{V\in \boldsymbol{V}}\left[\max\left[I(X,V), I(Y,V)\right]\right].
\end{split}
\end{equation*}
\label{asm:clust}
\end{asm}

\vskip -0.33in
This assumption is derived as follows.
Let $X \indep Y | \boldsymbol{Z}$, where $\boldsymbol{Z}$ is a minimal separating set. For a constraint-based causal discovery algorithm to identify this independence, it is essential that every $Z\in\boldsymbol{Z}$ is in the same cluster that includes $X$ and $Y$. To ensure this, every $Z\in\boldsymbol{Z}$ should have a correlation level with $X$ or $Y$, at least as the correlation level between $X$ and $Y$. That is, $\forall Z\in\boldsymbol{Z}, $ $I(Z,X)>I(X,Y)$ or $I(Z,Y)>I(X,Y)$.
Thus, if $X$ and $Y$ are in the same cluster, $\boldsymbol{Z}$ is also in that cluster. This is formally expressed by the first relation of \asmref{asm:clust}: $\min_{Z\in\boldsymbol{Z}}[\max[I(X,Z), I(Y,Z)]]\geq I(X,Y)$, where $\min_{Z\in\boldsymbol{Z}}$ essentially represents ``$\forall Z\in\mathbf{Z}$''. 
The second relation in \asmref{asm:clust} is $I(X,Y)>\min_{V\in\boldsymbol{V}}[\max[I(X,V), I(Y,V)]]$, where $\boldsymbol{V}$ is a set that does not include any $Z\in\boldsymbol{Z}$, $X$, and, $Y$. This relation assumes that the variables can be clustered. If no such redundant set, $\boldsymbol{V}$, exists it means that every variable in $\boldsymbol{X} \setminus (\{X,Y\}\cup\mathbf{Z})$ will have a stronger correlation with $X$ or $Y$ than the correlation between $X$ and $Y$. Thus, if $X$ and $Y$ are in the same cluster, then all other variables will be in the same cluster as well. 

\asmref{asm:clust} is required for achieving efficiency\footnote{Soundness and completeness of the method described in this paper does not rely on this assumption.} in the number of CI tests, and balances between: 1) allowing minimal separating sets to be discovered by $\mathrm{Alg}$ applied to a cluster, and 2) partitioning the variables into clusters.

\subsection{Domain Variable Clustering}\label{sec:spectral-clustering}

We now derive a clustering approach that complies with \asmref{asm:clust}.
Consider a fully connected undirected graph $\mathcal{U}$ over the domain variables $\boldsymbol{X}$. A symmetric similarity matrix $\boldsymbol{W}$ represents the weights of the edges in $\mathcal{U}$. The value of $\boldsymbol{W}_{i,j}$ is the weight of the edge between nodes $X_i,X_j\in\boldsymbol{X}$ and represents the correlation ``strength'' between these variables. The weight is the statistical measure of correlation, denoted $I$,  and calculated by the statistical independence test that is used by the baseline causal discovery algorithm. For example, mutual information for discrete variables and correlation coefficient for continuous variables (with a rapid density estimation, e.g., using \cite{gurwicz2004rapid}). Clustering can then be viewed as partitioning $\mathcal{U}$ into disjoint sub-graphs $\mathcal{U}_1,\ldots,\mathcal{U}_k$ by removing edges connecting the sub-graphs, where a cluster $\boldsymbol{X}_i$ consists of the nodes in sub-graph $\mathcal{U}_i$.
Partitioning $\mathcal{U}$ by minimizing the sum of weights of removed edges violates \asmref{asm:clust}, as discussed later. Moreover, as this sum increases with the number of removed edges, clustering algorithms based on this criterion favor creating small clusters of isolated nodes \citep{wu1993optimal}. As a solution, we follow \citet{shi2000normalized} that proposed the $k$-way  normalized cut (Ncut),
\begin{multline}\label{eq:ncut}
    \mathrm{Ncut}(\{\boldsymbol{X}_1,\ldots,\boldsymbol{X}_k\}) =  \\ = \sum_{i=1}^{k-1}\sum_{j=k+1}^k \nicefrac{\mathrm{cut}(\boldsymbol{X}_i, \boldsymbol{X}_j)}{\mathrm{assoc}(\boldsymbol{X}_i, \boldsymbol{X})},
\end{multline}
where $\mathrm{assoc}(\boldsymbol{X}_i, \boldsymbol{X})$ is the sum of weights of edges connecting each node in cluster $i$ to every other node in $\boldsymbol{X}$, and $\mathrm{cut}(\boldsymbol{X}_i, \boldsymbol{X}_j)$ is the sum of weights of edges connecting each node in cluster $i$ to every node in cluster $j$. 

This criterion complies with \asmref{asm:clust}. Let $X \indep Y | \boldsymbol{Z}$ where $\boldsymbol{Z}$ is a minimal separating set. Now, consider an undesired clustering: $\boldsymbol{X_1}=\{X,Y\}$ and $\boldsymbol{X_2}=\boldsymbol{Z}$. Then, $\mathrm{Ncut}(\boldsymbol{X}_1,\boldsymbol{X}_2) = \nicefrac{I(X,Z) + I(Y,Z)}{I(X,Z) + I(Y,Z) + I(X,Y)}$. To avoid such clustering, it is required to maximize the $\mathrm{Ncut}$ value $\forall Z\in\mathbf{Z}$. It is easy to see that this value is greater when $I(X,Z) > I(X,Y)$ than the value when $I(X,Z) < I(X,Y)$ and similarly for $I(Y,Z) > I(X,Y)$. Thus, this criterion complies with \asmref{asm:clust}. It is important to note that a criterion equal to the numerator of \eqref{eq:ncut} does not support \asmref{asm:clust}, as it ignores $I(X,Y)$.
In addition, $\mathrm{Ncut}$ diminishes the creation of small clusters. In fact, in the extreme case of  equal weights for all edges, $\mathrm{Ncut}$ is minimized for clusters with equal sizes.

\citet{shi2000normalized} showed that minimizing 2-way Ncut is equivalent to
\begin{equation}\label{eq:lapmincut}
    \min_u \nicefrac{(u^{\mathrm{T}}\boldsymbol{L}u)}{(u^{\mathrm{T}}\boldsymbol{D}u)} \qquad \mathrm{s.~t.}\quad u^{\mathrm{T}}\boldsymbol{D}1=0,
\end{equation}
\vskip -0.1in
where $u$ is an indicator vector of length $n$, $\boldsymbol{D}$ is a diagonal matrix with elements $\boldsymbol{D}_{i,i}=\sum_{j=1}^n \boldsymbol{W}_{i,j}$, and $\boldsymbol{L}=\boldsymbol{D}-\boldsymbol{W}$ is the Laplacian matrix. In our case, we can relax $u$ to take on real values, and the criterion can be minimized by solving the generalized eigenvalue system, $(\boldsymbol{D}-\boldsymbol{W})u=\lambda \boldsymbol{D}u$.
Taking the eigenvector corresponding to the smallest non-zero eigenvalue minimizes 
\begin{equation}\label{eq:rel-dist}
    \nicefrac{(\sum_{i,j} \boldsymbol{W}_{i,j}(u_i-u_j)^2)}{(\sum_i \boldsymbol{D}_{i,i}u_i^2)}.
\end{equation}
A Laplacian eigenmap \citep{belkin2003laplacian} is formed by concatenating the $m$ eigenvectors corresponding to the lowest non-zeros eigenvalues, $\boldsymbol{\Tilde{u}}=[u^1,\ldots,u^m]$. Thus, each domain variable $X_i\in\boldsymbol{X}$ is represented by a point $\boldsymbol{\Tilde{u}}_{(i,\cdot)}$ in $\mathbb{R}^m$. For our task, from \eqref{eq:rel-dist}, variables that are strongly correlated, \emph{relatively} to other pairs, will have a relatively small Euclidean distance $\mathbb{R}^m$. Finally, points $\Tilde{\boldsymbol{u}}$, representing variables $\boldsymbol{X}$ in $\mathbb{R}^m$, are clustered using k-means++ \citep{arthur2006k}. This procedure is known as spectral clustering.

\subsection{Proposed Method}

We consider the problem of learning a causal model, given a dataset for $n$ domain variables, $\{X_i\}^n_{i=1}$. Our method is composed of two main stages, commencing with a top-down hierarchical clustering stage followed by a bottom-up causal discovery in the backtracking stage. 

In the first stage, hierarchical clustering aims to alleviate the curse-of-dimensionality by partitioning the variable set into clusters, each of which potentially contains variables that are statistically related to each other to a large extent, thereby avoiding spurious connectivity to weaker and undesirable variables (\asmref{asm:clust}). 

Our method starts off by clustering the entire variable set from the outset into a number of clusters (see \secref{sec:spectral-clustering}), and thereafter successively clusters each of the resultant clusters furthermore, independently of the other clusters. This successive independent clustering process continues for each sub-cluster recursively, forming a tree of clusters, until a separability condition is met (explained later), at which point the entire variable set is clustered to subsets of variables. \figref{fig:method_illustration}(a) illustrates this process. We postulate that each such variable set has a high probability to manifest some structural motif \citep{milo2002network, Yang2018Learning}. A separability condition is used to determine the termination of the hierarchical clustering, and for that the eigenvalues of the graph’s Laplacian are used (\eqref{eq:lapmincut}). Generally, those close to zero correspond to isolated groups in the graph, and therefore if more than one such eigenvalue exists then the variable set of the sub-cluster is likely to contain more (relatively) disjoint groups within it, hence the hierarchical clustering process continues. In this case, the number of clusters for the next recursion call is the number of Laplacian’s eigenvalues that are close to zero ($k'$ in Algorithm \hyperlink{HCPCALG}{1}, line 9). Experimentally, it was observed that this criterion mostly terminates the clustering at optimal points. 

In the second stage, a bottom-up causal discovery algorithm, denoted $\mathrm{Alg}$, is applied to the sub-clusters, starting from the leaves of the cluster tree, and moving upwards towards the root of the tree. $\mathrm{Alg}$ is applied to the variable set of each sub-cluster independently to the other sub-clusters, assuming that being secluded and isolated by other irrelevant variables from other sub-clusters, is more probable to learn graphical models with reliable edges, i.e. with higher degree of certainty. In this paper we use the PC algorithm as $\mathrm{Alg}$. Even though the PC algorithm was chosen as the baseline algorithm, other constraint-based causal discovery methods \citep{tsamardinos2006max, rohekar2018bayesian, colombo2012learning} may be used and arguably improved, since this stage poses no assumptions or restrictions on the elements of any prospect method. 

After learning a graph for each sub-cluster, it is thereafter represented as a sub-graph. Further on, adjacent sub-graphs (those belonging to the same parent cluster) are backtracked in tandem upwards to their parent cluster, at which point they are merged as a single unified variable set. For that, edges are added between every node in one sub-cluster to every node in the other sub-clusters, and a list, $\mathcal{E}$, is formed from these added edges. That is, $\mathcal{E}$ lists the edges of bipartite graphs between every pair of sub-clusters. $\mathrm{Alg}$ is applied to the unified variable set and only the edges listed in $\mathcal{E}$ are tested for removal. That is, $\mathrm{Alg}$ does not consider new connections or removal of edges between any pair of variables within each sub-cluster. Ultimately, each sub-graph keeps its intra-cluster connectivity, presumably stemming from a more reliable variable set, and appends new inter-cluster connectivity, which were not taken into consideration at the former stage. Then, for preserving the completeness of $\mathrm{Alg}$, we apply $\mathrm{Alg}$ again to the unified variable set, this time considering all the remaining edges. The above process continues upwards the clusters tree and terminates after been applied at the root, at which point the final graph is formed from the entire variable set.

Note that $\mathrm{Alg}$ is required to learn about the edges in list $\mathcal{E}$. For the case of $\mathrm{Alg}=$PC-algorithm, we set the conditional-independence test function to return a result only for edges in $\mathcal{E}$. For edges not in this list, the function simply returns the existence or absence of the edge in the current graph as ``dependent'' or ``independent'' respectfully.

The main purpose of our approach is to improve the accuracy and efficiency of a given baseline algorithm by reducing the number of (unique) statistical tests, while maintaining the soundness and completeness of the baseline algorithm. Although we run the baseline algorithm on all the clusters in the backtracking phase, this inclusion will not undo any advantages of the clustering in terms of efficiency and accuracy. The reason for this is that each of the clusters is effectively: (a) containing only part of the nodes, so many conditional tests are avoided, and (b) sparser, since some edges were already removed and would not be tested anymore, and (c) condition tests that were already applied in previous steps would not be reapplied. So in essence, effectively we only avoid applying unnecessary condition tests, and consequently improve both speed and accuracy.

An improvement in performance of our method over a baseline is expected when \asmref{asm:clust} is complied, and this improvement is maximal when the sizes of the clusters are equal. As more clusters are revealed (while preserving causal sufficiency) the greater the improvement.

An additional virtue of our method is parallelism, that can be applied to successive independent clustering during the top-down stage, as well as to the causal discovery in independent sub-clusters during the bottom-up stage. The method is illustrated in \figref{fig:method_illustration}(a), and presented as Algorithm \hyperlink{HCPCALG}{1}. \figref{fig:method_illustration}(b) exemplifies the top-down 2-way clustering of the ALARM dataset.

\begin{thm}{
Let $\mathrm{Alg}$ be a causal discovery algorithm that take as input an initial graph and a list of edges to be learned. If $\mathrm{Alg}$ is sound and complete, then Algorithm \hyperlink{HCPCALG}{1} is sound and complete.
}
\end{thm}
\begin{proof}
\vskip -0.15in
The proof does not rely on \asmref{asm:clust} (applies to arbitrary partitions). The proof is given in Appendix A.
\vskip -0.3in
\end{proof}

Completeness of our approach is achieved by calling the sound and complete algorithm $\mathrm{Alg}(\boldsymbol{X}, \mathrm{edges}(\mathcal{G}_{\boldsymbol{X}}))$ for refining the result $\mathcal{G}_{\boldsymbol{X}}$ of the merged cluster (Algorithm \hyperlink{HCPCALG}{1}, line 18). In this call, all the graph edges are considered for learning, allowing undetected independence-relations withing the clusters to be detected.

\hypertarget{HCPCALG}{}
\begin{figure}[tb]
    \includegraphics[width=0.483\textwidth]{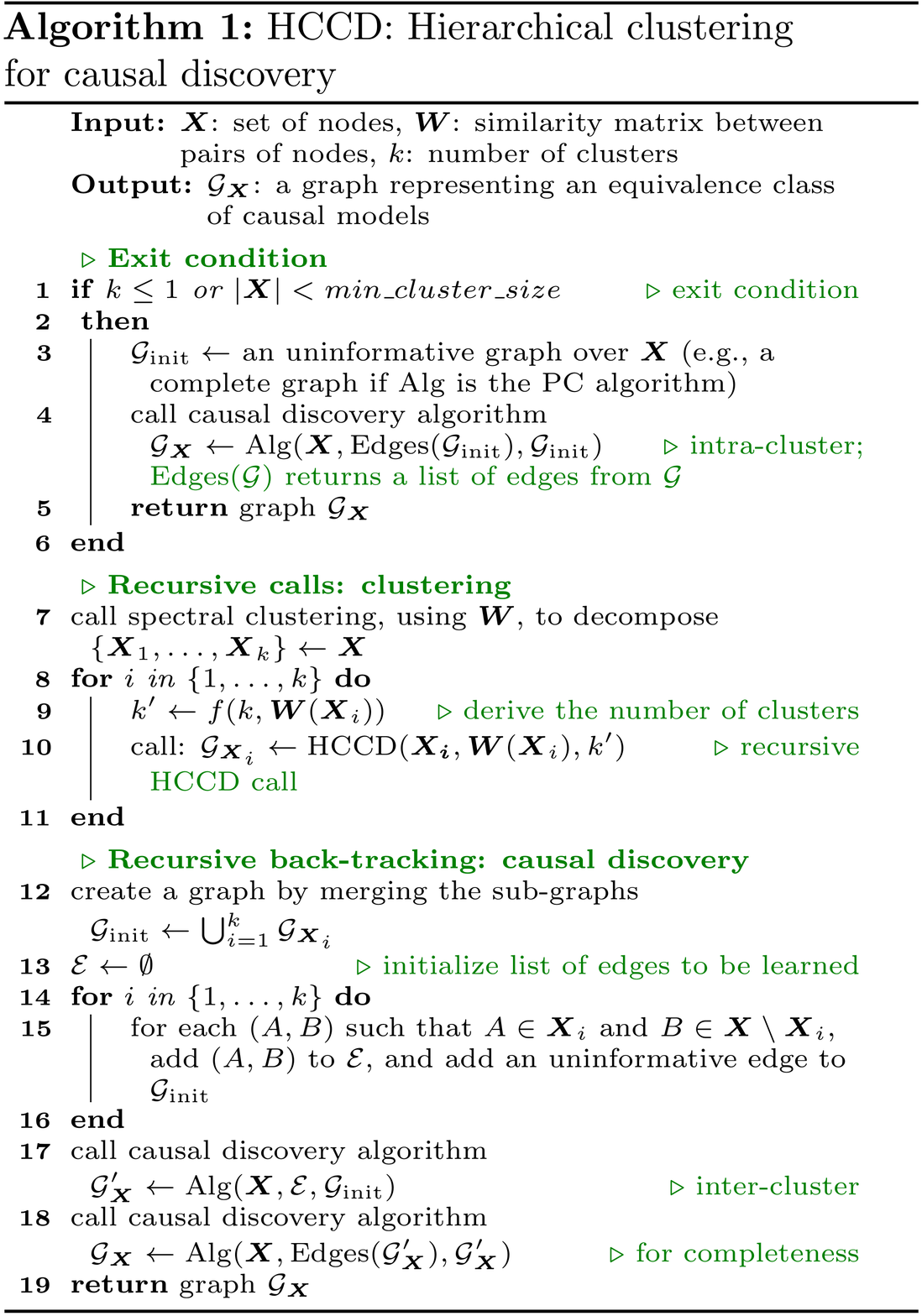}
    \vskip -0.25in
\end{figure}

\section{Experiments}

First, we evaluate several aspects of the HCCD wrapper using synthetically generated data.
The process we used for generating the data is detailed in Appendix B1. In addition, in appendix B2 we examine the gain achieved by the recursion, and the effect that the completeness requirement has on the accuracy.
Then, we evaluate qualitative measures of graphs learned using publicly available datasets.
In all our experiments, $\mathrm{Alg}$ is PC, a sound and complete algorithm \citep{spirtes2000}. Although it relies on the causal sufficiency assumption, it is often used as a first step of causal discovery in the presence of latent confounders and selection bias \citep{spirtes2000, claassen2013learning,colombo2012learning}.

\subsection{An Analysis using Synthetic Data}

In this section we evaluate the performance of the HCCD with respect to the number of training samples, and to the number of nodes in the graphs, compared to the baseline method (PC). For that, we measure the behaviour of 5 key aspects: 3 metrics of structural correctness - SID score \citep{peters2015structural}, structural Hamming distance (SHD), and causal accuracy. In addition, we measure the number of CI tests, and the run-time of the method, including the clustering.

\begin{figure}[ht!]
    \centering
    \subfigure[]{\includegraphics[width=0.22\textwidth]{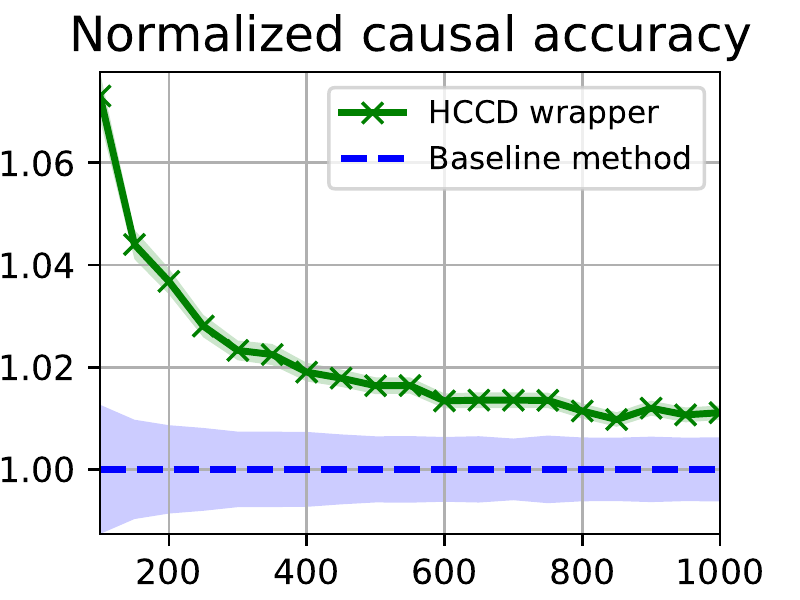}} 
    \subfigure[]{\includegraphics[width=0.22\textwidth]{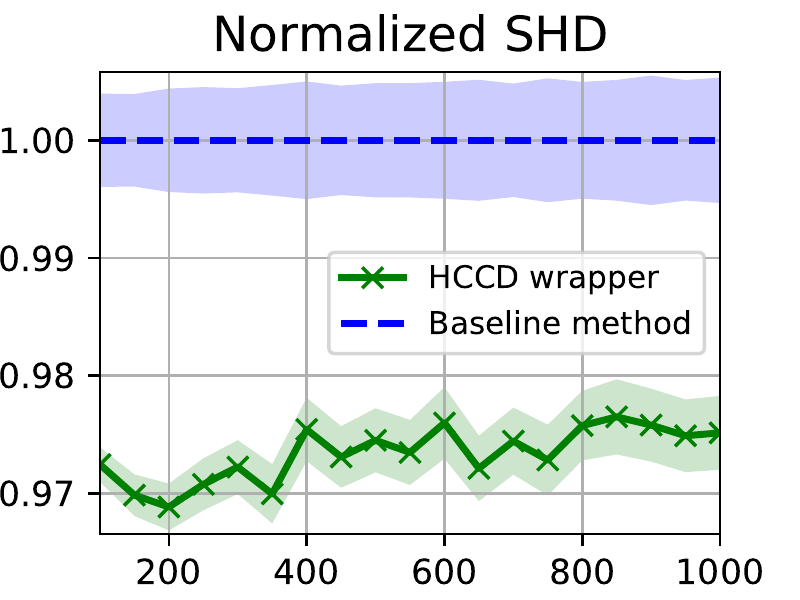}}
    \subfigure[]{\includegraphics[width=0.22\textwidth]{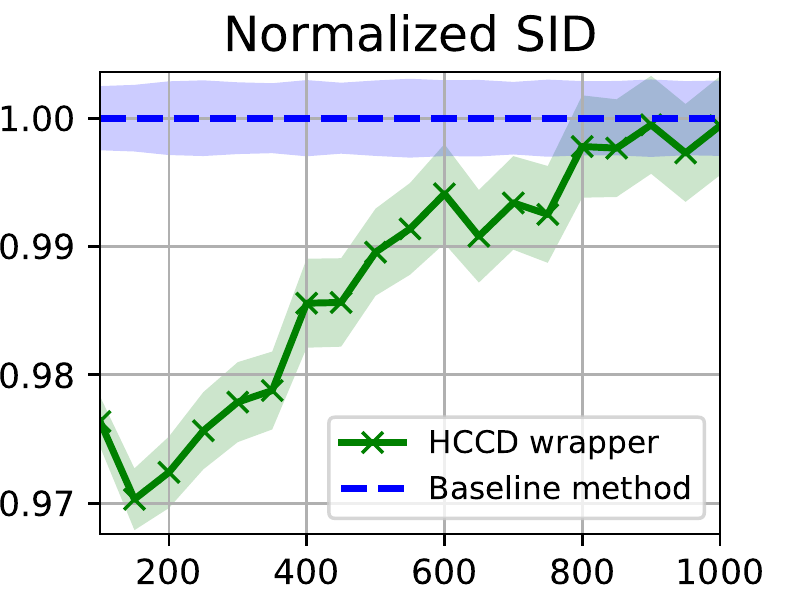}}
    \subfigure[]{\includegraphics[width=0.22\textwidth]{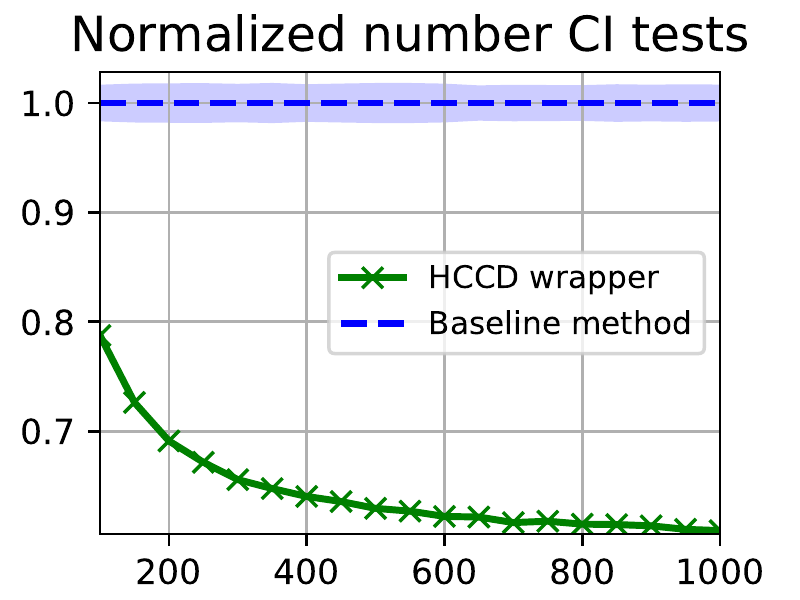}}
    \subfigure[]{\includegraphics[width=0.22\textwidth]{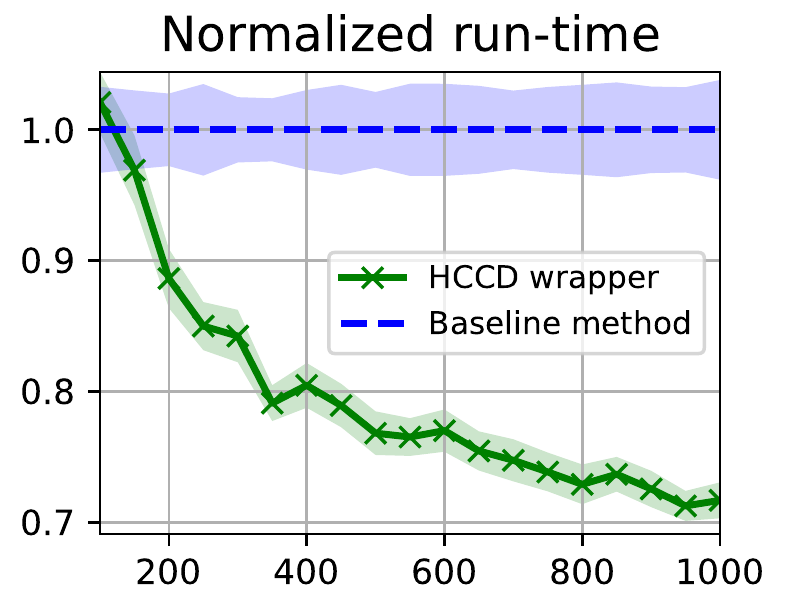}}
    \vskip -0.05in
    \caption{Performance of the HCCD wrapper, relatively to the baseline (PC), as a function of the number of training samples, for $100$ graph nodes. Values are average over 500 DAGs, and normalized by the PC score. (a) Causal accuracy (higher is better); (b) SHD (lower is better); (c) SID (lower is better); (d) Number of CI tests (lower is better); (e) Run-time (lower is better). The HCCD wrapper achieves improvements in all the metrics.}
    \vskip -0.15in
    \label{fig:scalability_to_training_set_100}
\end{figure}

\begin{figure}[h!]
    \centering
    \subfigure[]{\includegraphics[width=0.22\textwidth]{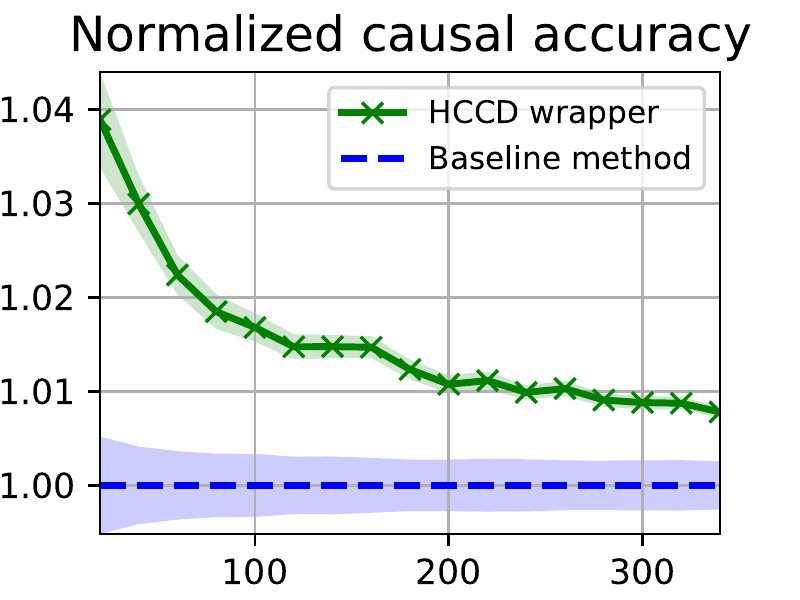}} 
    \subfigure[]{\includegraphics[width=0.22\textwidth]{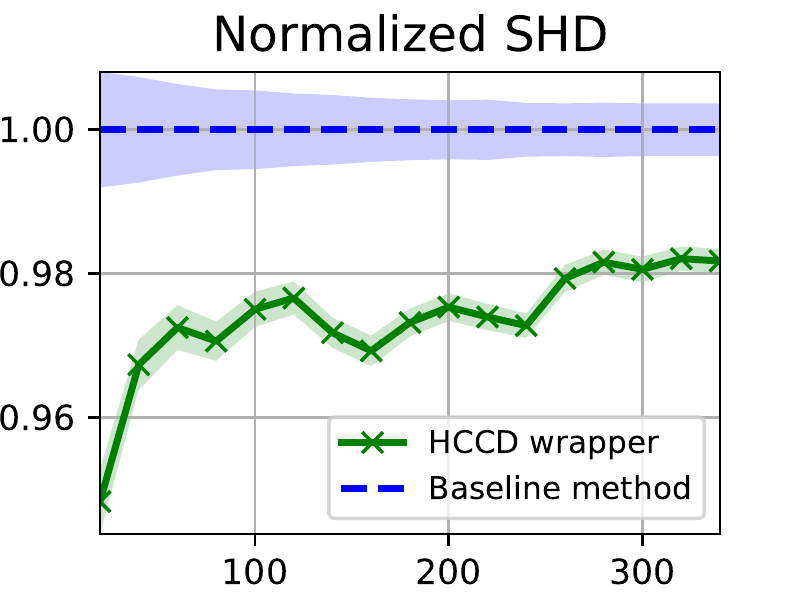}}
    \subfigure[]{\includegraphics[width=0.22\textwidth]{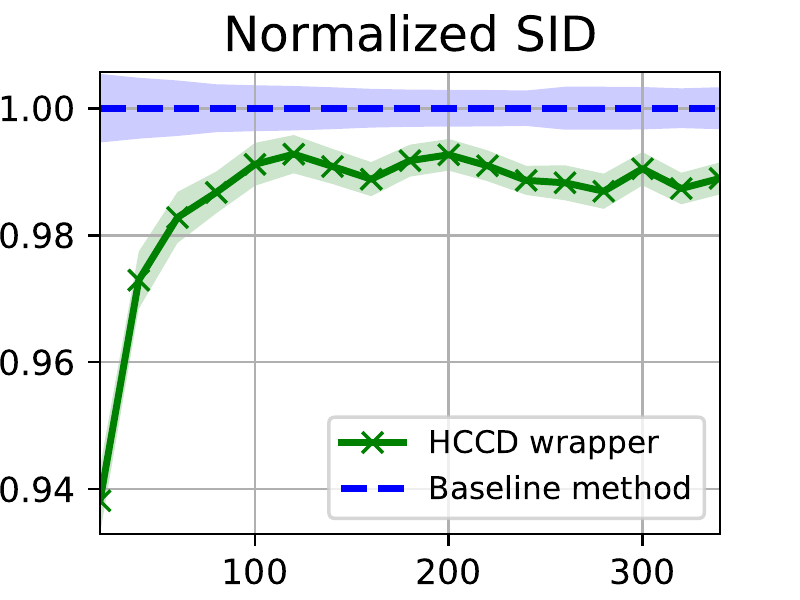}}
    \subfigure[]{\includegraphics[width=0.22\textwidth]{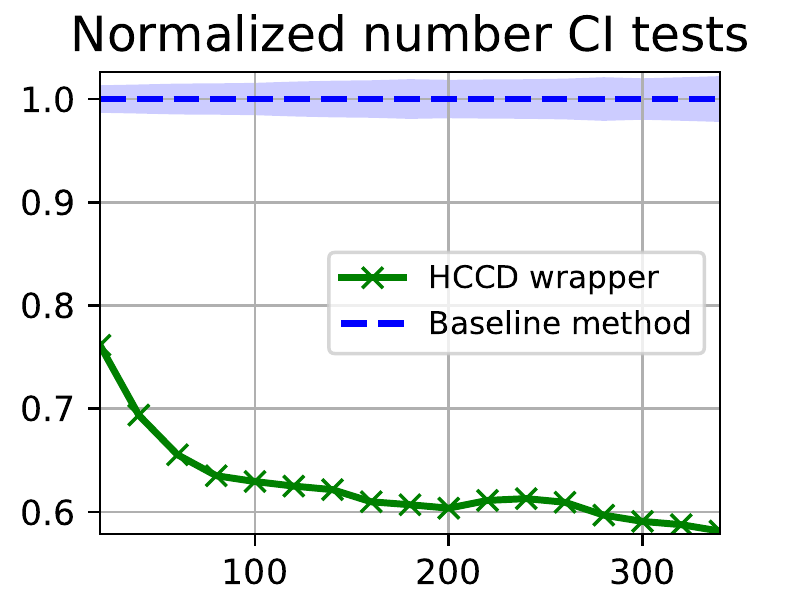}}
    \subfigure[]{\includegraphics[width=0.22\textwidth]{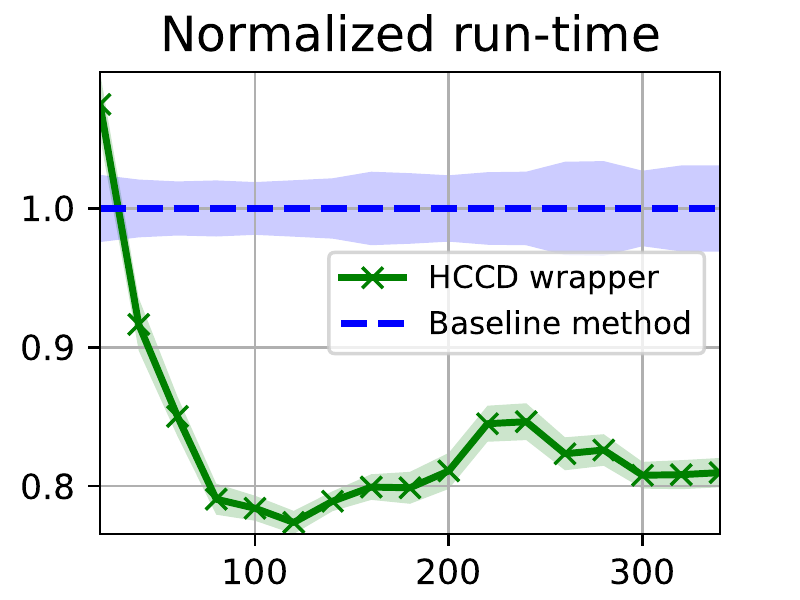}}
    \vskip -0.07in
    \caption{Performance of the HCCD wrapper, relatively to the baseline (PC), as a function of the number of graph nodes, for $500$ training samples. Values are average over $500$ DAGs, and normalized by the PC score. (a) Causal accuracy (higher is better); (b) SHD (lower is better); (c) SID (lower is better); (d) Number of CI tests (lower is better); (e) Run-time (lower is better). The HCCD wrapper achieves improvements in all the metrics.}
    \vskip -0.23in
    \label{fig:scalability_to_number_of_nodes}
\end{figure}

\figref{fig:scalability_to_training_set_100} shows the performance of the HCCD wrapper with respect to the number of training samples, for graphs with $n=100$ nodes. The figures show mean $\pm$ std of $500$ independent tests (DAGs), and values are normalized by the PC score in order to visualize the improvement over the baseline method. Additional experiments, for $n\in\{20,50,200,1000\}$, are presented in Appendix C. It is evident that the HCCD wrapper is superior to the baseline for all 3 structural correctness metrics along the entire range of the training set size and for every $n$. In addition, there is an evident saving in the number of CI tests, and importantly in run-time (includes the clustering stage) along the entire range of the training set size. One exception is for the case of $n=20$, for which the HCCD run-time is higher. This is expected since the run-time overhead of the clustering stage overtakes the saving in run-time gained by using fewer statistical tests in datasets with a small number of nodes. Nevertheless, for the common real-world cases, datasets having many variables, the HCCD wrapper achieves a significant reduction in run-time. Additionally, it is evident that the run-time reduction increases with the increase of the number of training samples, i.e. larger training sets benefit from a greater decrease in run-time.

\figref{fig:scalability_to_number_of_nodes} shows the performance of the HCCD wrapper with respect to $n$, the number of nodes, for $500$ training samples. The figures show mean $\pm$ std of 500 independent tests (DAGs), and values are normalized by the PC score in order to analyze the improvement over the baseline method. It is evident that the HCCD wrapper is superior to the baseline for all 3 structural correctness metrics along the entire range of $n$. Moreover, saving in number of CI tests is evident, and importantly a reduction in run-time (which includes the clustering stage) for the entire range of $n$.

\subsection{Real-World Data}
\begin{figure}[h!]
    \centering
        \includegraphics[width=0.48\textwidth]{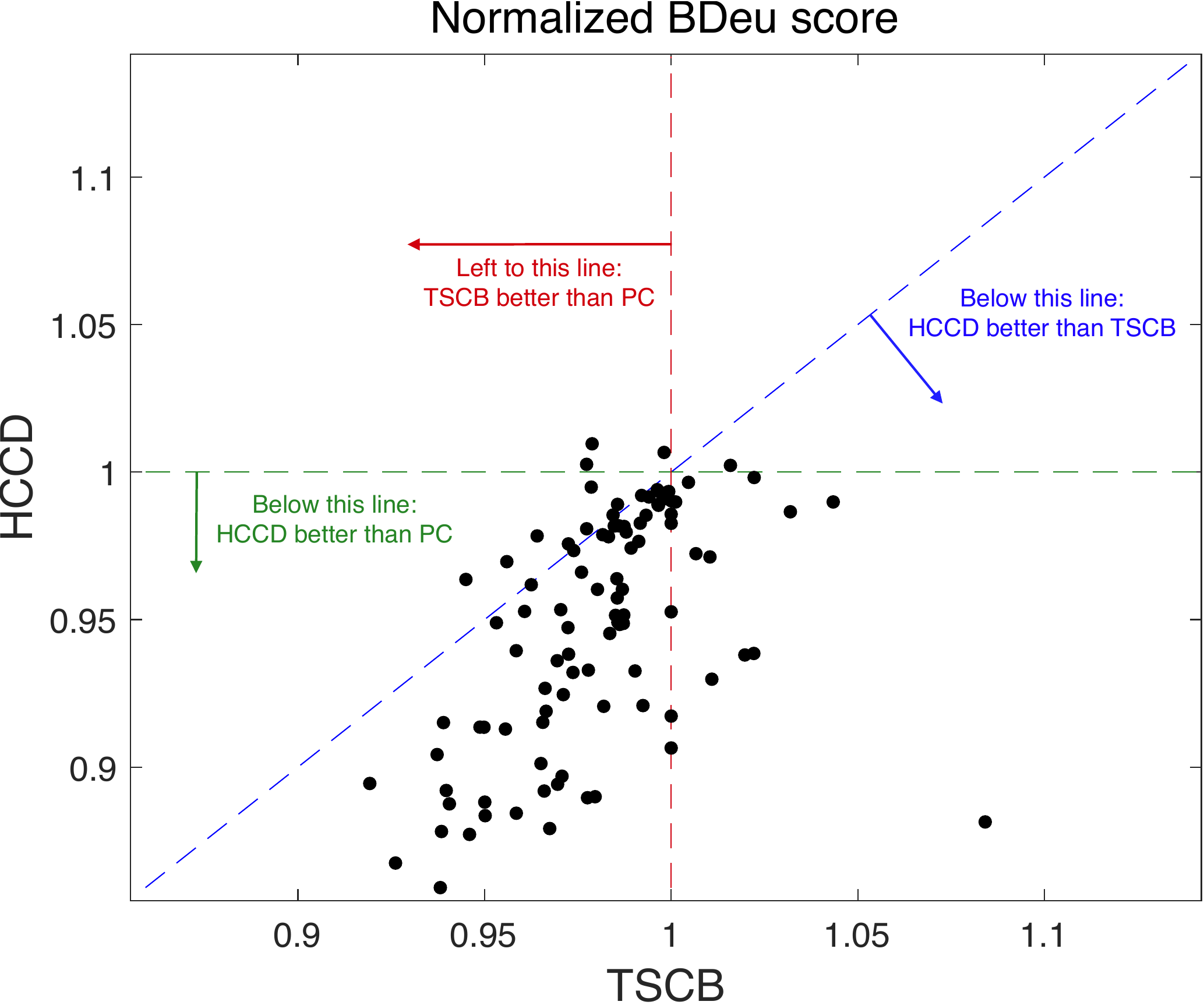} 
        \vskip -0.05in
        \caption{Scatter plot of normalized BDeu score, comparing HCCD, TSCB, and PC, evaluated on the 11 datasets from \tabref{table_BDeu_Scores}, each consists of 10 different training and test sets. The scores are normalized by the PC BDeu score, and so lower is better. Points below the green dashed line correspond to better results of HCCD compared to the PC, which are 97\% of the cases. Points to the left of the red dashed line correspond to better results of TSCB compared to PC, which are 82\% of the cases. Points below the blue dashed line correspond to better results of HCCD compared to TSCB, which are 90\% of the cases.}
        \label{fig:scatter_plot_score}
\end{figure}

In this section we evaluate and compare the accuracy of our method over 10 publicly available datasets from the bnlearn package \citep{marco2010bnlearn}, and 1 dataset from the Neuropathic Pain Diagnosis Simulator \citep{rubio2019pain}, all of which represent real decision support systems that cover a wide range of real life applications, such as medicine, agriculture, weather forecasting, financial modeling and animal breeding. Each of those datasets consists of 10 training sets, having 500 samples each, and corresponding 10 separate test sets, having 5000 samples each, for evaluating several qualitative measures of structural correctness. Thus, for each of the 11 datasets, the experiments were repeated 10 times. The number of domain variables across the datasets spans from tens to hundreds.
\vskip 0.1in
The first metric we measure is the BDeu score \citep{chickering1995learning}, which under certain assumptions corresponds to the posterior probability of the learned graph. \citet{tsamardinos2006max} noted that this score does not rely on the true graph and may not be related to it, as it is not known in practice to what extent its underlying assumptions hold (e.g., Dirichlet distribution of the hyper-parameters). Nevertheless, since this score does not require knowing the true graph, it has a great value in practical situations. Moreover, this score is often used to tune the baseline parameters \citep{yehezkel2009rai}. \figref{fig:scatter_plot_score} shows a scatter plot of normalized BDeu score, comparing HCCD, TSCB, and PC, evaluated on the 11 datasets, each consists of the 10 different training and test sets (total of 100 points). The BDeu scores are normalized by the PC BDeu score, and so a lower normalized score is better. In 97\% of the cases, HCCD is better than PC. In 82\% of the cases, TSCB is better than PC. Lastly, in 90\% of the cases, HCCD is better than TSCB. As evident from the figure, HCCD is superior to the other methods. Additionally, for each complete dataset, the mean $\pm$ std BDeu score (unnormalized) is presented in \tabref{table_BDeu_Scores}, and better results for the HCCD are observed on all the datasets. 
\vskip 0.01in

\begin{table}[h!]
\caption{BDeu Scores (higher is better) of PC, TSCB, and HCCD on various datasets.}
\vskip -0.2in
\label{table_BDeu_Scores}
\begin{center}
\begin{tiny}
\begin{sc}
\begin{tabular}{lcccr}
\toprule
Data set & PC & TSCB & HCCD \\
\midrule
Alarm       & -60290    $\pm$ 2750  & -57920    $\pm$ 1280& \textbf{-55852}$\pm$ 1703 \\
Child       & -67309    $\pm$ 1059  & -66554    $\pm$ 765&  \textbf{-64539}$\pm$ 290 \\
Insurance   & -74690    $\pm$ 1543  & -73848    $\pm$ 1315& \textbf{-73469}$\pm$ 1038 \\
Mildew      & -293679   $\pm$ 11072 & -290456   $\pm$ 14157 & \textbf{-267266}$\pm$ 3858 \\
Hailfinder  & -301499   $\pm$ 2309  & -292020   $\pm$ 3365 &  \textbf{-290200}$\pm$ 3503 \\
Barley      & -358461   $\pm$ 3592  & -352608   $\pm$ 6365 &  \textbf{-350807}$\pm$ 3419 \\
Munin       & -451571   $\pm$ 2686  & -434700   $\pm$ 4394 &  \textbf{-401007}$\pm$ 3543 \\
WIN95PTS    & -64439   $\pm$ 768    & -62990    $\pm$ 947  &  \textbf{-60807}$\pm$ 876 \\
PathFinder  & -274163  $\pm$ 2334   & -262620   $\pm$ 5004 &  \textbf{-248600}$\pm$ 2822 \\
Hepar2      & -168822  $\pm$ 582   & -168528   $\pm$ 549 &  \textbf{-167489}$\pm$ 533 \\
NeuroPain & -185340  $\pm$ 678   & -182572   $\pm$ 1265 &  \textbf{-181670}$\pm$ 538 \\

\bottomrule
\end{tabular}
\end{sc}
\end{tiny}
\end{center}
\vskip -0.3in
\end{table}

\begin{figure}[h!]
        \centering
        \includegraphics[width=0.48\textwidth]{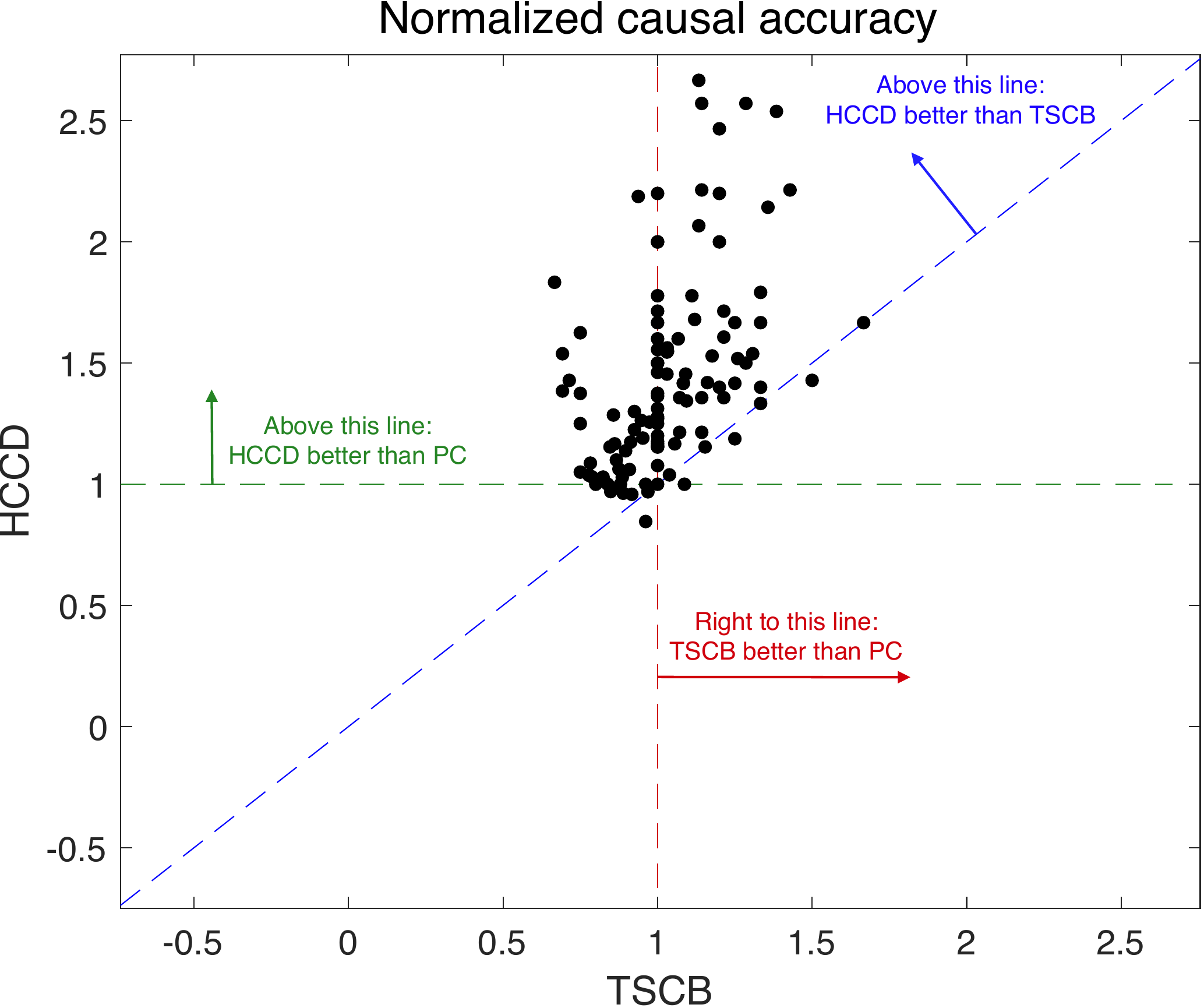} 
        \vskip -0.05in
        \caption{Scatter plot of normalized causal accuracy, comparing HCCD, TSCB, and PC, evaluated on the 11 datasets from \tabref{tab:shd_causal_acc}, each consists of 10 different training sets. The values are normalized by the PC value, and higher is better. Points above the green dashed line correspond to better results of HCCD compared to the PC, which are 91\% of the cases. Points to the right of the red dashed line correspond to better results of TSCB compared to PC, which are 51\% of the cases. Lastly, points above the blue dashed line correspond to better results of HCCD compared to TSCB, which are 93\% of the cases.}
        \label{fig:scatter_plot_acc}
    \vskip -0.2in
\end{figure}

We also calculate causal accuracy \citep{claassen2012bayesian} as an evaluation metric to causal discovery. \figref{fig:scatter_plot_acc} shows a scatter plot of causal accuracy, comparing HCCD, TSCB, and PC, evaluated on the 11 datasets, each consists of 10 different training sets (total of 100 points). The causal accuracies are normalized by the PC causal accuracy, and so higher is better. In 91\% of the cases, HCCD is better than PC. In 51\% of the cases, TSCB is better than PC. Lastly, in 93\% of the cases, HCCD is better than TSCB. As evident from the figure, HCCD is superior to the other methods. Additionally, for each complete dataset, the mean $\pm$ std causal accuracy is presented in \tabref{tab:shd_causal_acc}, and better results for the HCCD are observed on all the datasets, demonstrating improved ability to recover the ground-truth causal graph.

\begin{table}[h!]
\caption{SHD (lower is better) and causal accuracy (higher is better) comparison for various datasets.}
\vskip -0.2in
\label{tab:shd_causal_acc}
\begin{center}
\begin{tiny}
\begin{sc}
\begin{tabular}{lccc}
\toprule
 & \multicolumn{3}{c}{Structural Hamming distance} \\
Data set & PC & TSCB & HCCD \\
\midrule
Alarm       & 30.50$\pm$ 3.14   & 39.6$\pm$ 4.55 & \textbf{28.80}$\pm$ 3.55    \\
Child       & 18.50$\pm$ 1.27   & 19.4$\pm$ 1.35 & \textbf{18.0}$\pm$ 1.25  \\
Insurance   & 42.90$\pm$ 2.62   & 46.40$\pm$ 4.27 & \textbf{42.20}$\pm$ 2.84   \\
Mildew      & 46.80$\pm$ 1.62   & 46.80$\pm$ 2.04 & \textbf{45.70}$\pm$ 0.95  \\
Hailfinder  & 80.30$\pm$ 2.63   & 86.20$\pm$ 2.49 & \textbf{80.10}$\pm$ 2.03   \\
Barley      & 83.90$\pm$ 0.74   & 83.90$\pm$ 0.99 & \textbf{81.60}$\pm$ 1.26   \\
Munin       & 283$\pm$ 1.06     & 286.10$\pm$ 1.37 & \textbf{279.60}$\pm$ 2.60   \\
WIN95PTS    & 99.30$\pm$ 4.30   & 103.20$\pm$ 7.28 & \textbf{97.80}$\pm$ 7.06   \\
PathFinder  & \textbf{193.10}$\pm$ 0.94 & 199.30$\pm$ 2.21 & 195.20$\pm$ 2.56     \\
Hepar2      & 115.70$\pm$ 2.21& 117.80$\pm$ 0.92 & \textbf{114.20}$\pm$ 2.53    \\
NeuroPain   & 796.70$\pm$ 13.71& 804$\pm$ 23.04 & \textbf{791}$\pm$ 8.24      \\

\bottomrule
\end{tabular}
\end{sc}
\end{tiny}
\end{center}

\begin{center}
\begin{tiny}
\begin{sc}
\begin{tabular}{lccc}
\toprule
  & \multicolumn{3}{c}{Causal accuracy}\\
Data set & PC & TSCB & HCCD\\
\midrule
Alarm       &  0.700   $\pm$ 0.039 & 0.608   $\pm$ 0.044 &  \textbf{0.727}$\pm$ 0.036 \\
Child       &  0.440   $\pm$ 0.067 & 0.448   $\pm$ 0.064 &  \textbf{0.597}$\pm$ 0.044 \\
Insurance   &  0.476   $\pm$ 0.038 & 0.446   $\pm$ 0.044 &  \textbf{0.485}$\pm$ 0.038 \\
Mildew      &  0.143   $\pm$ 0.029 & 0.126   $\pm$ 0.022 &  \textbf{0.235}$\pm$ 0.019 \\
Hailfinder  &  0.056   $\pm$ 0.012 & 0.060   $\pm$ 0.010 &  \textbf{0.082}$\pm$ 0.010 \\
Barley      &  0.180   $\pm$ 0.027 & 0.203   $\pm$ 0.022 &  \textbf{0.233}$\pm$ 0.017 \\
Munin       &  0.051   $\pm$ 0.002 & 0.061   $\pm$ 0.005 &  \textbf{0.121}$\pm$ 0.011 \\
WIN95PTS    &  0.320   $\pm$ 0.030 & 0.327   $\pm$ 0.026 &  \textbf{0.441}$\pm$ 0.015 \\
PathFinder  &  0.066   $\pm$ 0.003 & 0.064   $\pm$ 0.011 &  \textbf{0.088}$\pm$ 0.008 \\
Hepar2      &  0.132   $\pm$ 0.024 & 0.139   $\pm$ 0.019 &  \textbf{0.181}$\pm$ 0.017 \\
NeuroPain   &  0.037   $\pm$ 0.005 & 0.042   $\pm$ 0.003 &  \textbf{0.057}$\pm$ 0.004 \\

\bottomrule
\end{tabular}
\end{sc}
\end{tiny}
\end{center}

\vskip -0.3in
\end{table}

\vskip 0.2in
In addition, we measure the structural hamming distance (SHD) between the learned graph and the ground-truth graph. SHD calculates the number of edge insertions, deletions or flips in order to transform one graph to another graph. For each of the 11 dataset, the mean $\pm$ std SHD is presented in \tabref{tab:shd_causal_acc}. For all the datasets except one, HCCD is better than the other methods.

\section{Conclusions}

We propose the HCCD wrapper for causal discovery algorithms (baseline algorithms). HCCD preserves soundness and completeness of the baseline algorithm, while reducing the number of statistical tests, increasing the accuracy of the resulting graph, and reducing the run-time. For constraint-based baseline algorithms, it is assumed that each pair of variables, not adjacent in the true underlying graph, is more strongly correlated to at least one of its separating sets than to other variables not in any of their separating set. Therefore, this property of relative strength of correlation is used by our method to hierarchically partition the domain variables, minimizing the number of independence relations that are not detectable from the cluster variables alone. 

Using synthetically generated graphs and data, and selectively limiting certain aspect of HCCD, we demonstrated that recursion and completeness-requirement greatly improve efficiency of the learning procedure and accuracy of the resulting causal graph. Applying our method to real-world graphs, and common publicly available datasets, we demonstrated that HCCD learns significantly more accurate graphs, compared to the PC baseline algorithm.

Finally, we conjecture that scoring-based algorithms may benefit from the HCCD wrapper as well by defining corresponding ``similarty'' measures. Thus, the search strategy is applied to smaller search spaces, independently and in parallel. We suspect that this will lead to avoiding local maximum and finding higher maximum points.

\bibliography{RCSL}

\appendix
\section{Proofs}

In the paper, we relate specifically to constraint-based causal discovery algorithms, relaying on the faithfulness and causal Markov assumption. Causal sufficiency assumption is not required. Let $\mathcal{G}$ be a causal DAG over a set of variables $\boldsymbol{V}=\boldsymbol{X}\cup\boldsymbol{H}\cup\boldsymbol{S}$, where $\boldsymbol{X}$, $\boldsymbol{H}$, and $\boldsymbol{S}$ are the observed, latent, and selection variable sets, respectively. 

Let $\mathrm{Alg}$ be a causal discovery algorithm, and $\mathrm{ClustCD}$ be the following procedure:
\begin{enumerate}
    \item Given observed data for domain variables $\boldsymbol{X}=\{A, B, \ldots\}$, partition $\boldsymbol{X}$ into $k$ disjoint subsets $\boldsymbol{X}_1, \ldots, \boldsymbol{X}_k$.
    \item Call $\mathrm{Alg}$ for each $\boldsymbol{X}_i$ (intra-cluster).
    \item Call $\mathrm{Alg}$ for edges between any pair $(A,B)$ such that $A\in\boldsymbol{X}_i$ and $B\in\boldsymbol{X}_j$, for all $i,j\in\{1,\ldots,k\}, i\neq j$ (inter-cluster).
\end{enumerate}

\begin{thm}\label{thm:twostep}

If $\mathrm{Alg}$ is a sound and complete causal discovery algorithm, then for any partition of $\boldsymbol{X}$ procedure $\mathrm{ClustCD}$ is sound for $\boldsymbol{X}$, but not complete.
\end{thm}

That is, given some partition of the variable set, $\boldsymbol{X}=\{\boldsymbol{X}_1, \ldots, \boldsymbol{X}_k$\}, a pair $A,B\in\boldsymbol{X}_i$, unconnected in the equivalence class of $\mathcal{G}$ (a CPDAG under causal efficiency, otherwise a PAG), might not be tested for independence in steps 2 and 3 of $\mathrm{ClustCD}$, conditioned on any of their separating sets.
Once a skeleton is constructed and separating sets are identified, additional independence tests are not required for orienting the edges (e.g., orientation rules \citep{zhang2008completeness}, LiNGam \citep{shimizu2006linear}, non-linear Gaussian models \citep{hoyer2009nonlinear}). Nevertheless, they rely on the completeness of skeleton learning. Thus, if not all detectable independence relations are identified, the orientation of the edges will not be complete as well.

\begin{proof}
By contradiction, $\mathrm{ClustCD}$ is complete. Then, it follows that for every partition of $\boldsymbol{X}$, every pair of unconnected nodes in the equivalence class of $\mathcal{G}$ will be tested for independence in steps 2 and/or 3, conditioned on at least one of their separating sets. A contrary example can be easily constructed. For example, $\mathcal{G}$ is $A\rightarrow D \leftarrow C \rightarrow E \leftarrow B$ and a clustering $\boldsymbol{X}_1=\{D,E\}, \boldsymbol{X}_2=\{A,B,C\}$. Since $C$ is the only separating set for $D,E$ and $C\notin\boldsymbol{X}_1$ it will not be included in independence tests for $\boldsymbol{X}_1$ in step 2 of $\mathrm{ClustCD}$. In step 3, since only nodes not in the same cluster will be tested, and $D,E$ are in the same cluster, they will not be tested for independence. Thus independence of $D,E$ conditioned on $C$ will not be tested.
\end{proof}

Note that \thmref{thm:twostep} refers to an arbitrary partition of $\boldsymbol{X}$. Nevertheless, one may consider a specific partition 
such that every independence relation within each cluster is detectable in step 2 of $\mathrm{ClustCD}$. That is, for every pair in a cluster, $\boldsymbol{X}_i$, that are unconnected in the equivalence class of $\mathcal{G}$, at least one separating set exists in the same cluster, $\boldsymbol{X}_i$. However, some graphs do not have such partition (assuming each cluster consists of at least one independence relation needed to be recovered). One example of such graph is $A\rightarrow \{B,C,D,E\}, \{B,C,D,E\}\rightarrow F$, and $\{B,C,D,E\}$ are unconnected ($A$ is their parent and $F$ is a collider). Thus, \thmref{thm:twostep} is permissible for these specific partitions as well (assuming each cluster contains at least one independence relation to be recovered).

\begin{thm}{
Let $\mathrm{Alg}$ be a causal discovery algorithm that takes as input an initial graph and a connectivity to be retained. If $\mathrm{Alg}$ is sound and complete, then Algorithm 1 is sound and complete.
}
\end{thm}

\begin{proof}

The proof follows immediately from line 18 of the Algorithm. The sound and complete causal discovery algorithm $\mathrm{Alg}$ is called, and every pair of nodes connected by an edge in $\mathcal{G}'_{\boldsymbol{X}}$ can be tested for independence conditioned on any subset of $\boldsymbol{X}$. 

Nevertheless, we also prove by mathematical induction for better clarity on how completeness is obtained gradually. We prove for recursion depth 1 (base case) and then for recursion depth $d+1$ (inductive step).

\underline{Base case}: Recursion depth is 1. Input variable set $\boldsymbol{X}$ is partitioned into $\{\boldsymbol{X}_1,\ldots,\boldsymbol{X}_k\}$ clusters. As recursion depth of 1 is assumed, any further recursive calls (line 10) comply with the exit condition (line 1). As a result $\mathrm{Alg}$ is called for learning a causal graph for each of the clusters independently and returned. Since $\mathrm{Alg}$ is sound and complete, every independence relation between nodes in a cluster $\boldsymbol{X}_i$ that are detectable by conditioning on some subset $\boldsymbol{Z}\subset\boldsymbol{X}_i$ will be detected (due to soundness, no false independencies will be returned).
$\mathrm{Alg}$ is called again, fixing (retaining) the sub-graph for each cluster, learning only inter cluster edges (line 17). However, the resulting graph $\mathcal{G}'_{\boldsymbol{X}}$ is not complete (\thmref{thm:twostep}). Calling $\mathrm{Alg}$ again, this time without fixing the sub-graphs, allows testing every pair, adjacent in $\mathcal{G}'_{\boldsymbol{X}}$ conditioned on any subset of $\boldsymbol{X}$. Thus, if $\mathrm{Alg}$ is complete, every independence relation in the equivalence class of $\mathcal{G}$, where at least one separating set is in $\boldsymbol{X}$, will be detected, and Algorithm 1 is complete.

\underline{Inductive step}: Assume that the algorithm is sound and complete for recursion depth $d$ and prove for recursion depth $d+1$. Input variable $\boldsymbol{X}$ is partitioned into $\{\boldsymbol{X}_1,\ldots,\boldsymbol{X}_k\}$ clusters. A recursive call (line 10) for cluster $\boldsymbol{X}_i$, and subsequent $d$ recursive calls, ensures that all independence relations, with separating sets within $\boldsymbol{X}_i$ will be detected. All independence relations between clusters are identified in line 17. Finally, $\mathrm{Alg}$ is called in line 18 for $\boldsymbol{X}$ using an initial graph $\mathcal{G}'_{\boldsymbol{X}}$, allowing every adjacent pair to be tested for independence on any subset of $\boldsymbol{X}$.
\end{proof}

\section{Synthetic Data Generation and Ablation Study}

In this section describe the process and present the results of Section 4 regarding the synthetic data generation and the ablation study.

\subsection{Synthetic Data Generation}
\vskip -0.1in
We generate random DAGs and sampled data using the following procedure. A DAG, having $n$ variables and connectivity factor $\rho$ is sampled in the following way. First, an adjacency matrix $\boldsymbol{A}$ of a DAG $\mathcal{G}$ is created by independent realization of $\mathrm{Bernoulli}(\nicefrac{\rho}{(n-1)})$ in the upper triangle. Importantly, if the resulting DAG is unconnected, we repeat until a connected DAG is sampled.
Then, a weight matrix $\boldsymbol{W}$ for the graph edges is created by sampling from $\mathrm{Uniform}([0.1, 1])$ for each non-zero element in $\boldsymbol{A}$.
Finally, graph $\mathcal{G}$ is treated as a statistical model by setting conditional probabilities $p(X_i|\mathrm{Pa}_{\boldsymbol{A}}(X_i))=\boldsymbol{W}_{(\cdot,i)}\boldsymbol{A}^{\mathrm{T}}_{(\cdot,i)}+\epsilon_i$, where $\epsilon_i\sim\mathcal{N}(0,1)$.

\subsection{Ablation Study}
\vskip -0.1in
We evaluate how the density of the underlying graph affects the number of independence tests performed. In addition, we examine the gain achieved by the recursion, and the effect that the completeness requirement has on the accuracy. 

Conditional independence is determined if partial correlation, after applying Fisher's z-transform, is equal to zero, with significance level $\alpha=0.01$. In each experiment 100 DAGs are created. For each DAG, the number of nodes is $n=100$, connectivity factor is $\rho\in\{3,4,5,6,7\}$, and the number of data samples $\ell=1000$. For HCCD, in Algorithm 1-line 9, $k'=k=2$. For this experiment, we limit to only one recursive call. Thus, 4 clusters are formed before backtracking. We compare this to case where no recursive calls are performed, achieved by setting $k=4$ and $k'=1$ in Algorithm 1-line 9, denoted ``\emph{HCCD-flat}'' (this is equivalent to the TSCB wrapper). We consider a modified version of HCCD where the second call, line 18 in Algorithm 1, is removed from the recursion and called only after the algorithm concludes (this modification still preserves completeness). We denote this modified version, ``\emph{HCCD-not-c}''.
From \tabref{tab:num_ci}, it is evident that the number of required independence tests is significantly reduced by HCCD wrapper for the PC algorithm. Moreover, when comparing to HCCD-flat (same number of clusters without recursion) it is evident that HCCD benefits from recursion. Finally, we can see that removing the completeness requirement for each cluster (HCCD-not-c) has constant improvement over PC but is inferior to HCCD. We attribute this to the many undetectable independence relations that are represented by retained edges ($\mathcal{L}$). This causes PC to consider more condition sets resulting in more independence tests. As in this experiment we limited HCCD to one recursive call and to minimal number of clusters, gain in accuracy, compared to PC or modified versions of HCCD, is statistically insignificant. Nevertheless, this same setting is adequate for evaluating the effect of the completeness requirement on the accuracy. We calculate the ratio of errors, dividing average number of HCCD errors by average number of errors of its non-complete modified version (without line 18, Algorithm 1). For DAG connectivity factors ($\rho$) $[3, 4, 5, 6, 7]$, extra-edges errors ratios are, $[0.11, 0.08, 0.1, 0.16, 0.21]$, and missing edges ratios are $[1.11, 1.12, 1.12, 1.08]$, respectively. 
The ratios of errors are reported in \tabref{tab:acc_completeness}. 
Although the completeness requirement results in some increase in missing edges, it also results in a significant decrease in extra edges.
\begin{table}[bth]
\vskip -0.1in
\caption{
Normalized number of independence tests required for recovering DAGs having various densities (parametrized by the connectivity factor $\rho$).
}
\vskip -0.15in
\label{tab:num_ci}
\begin{center}
\begin{scriptsize}
\begin{sc}
\begin{tabular}{lcccc}
\toprule
Density & PC & HCCD-flat & HCCD-not-c & HCCD\\
\midrule
$\rho=3$       & 1.00  & 0.89    & 0.86    & \textbf{0.80} \\
$\rho=4$       & 1.00  & 0.82    & 0.85    & \textbf{0.74} \\
$\rho=5$       & 1.00  & 0.78    & 0.86    & \textbf{0.66} \\
$\rho=6$       & 1.00  & 0.77    & 0.81    & \textbf{0.61} \\
$\rho=7$       & 1.00  & 0.76    & 0.74    & \textbf{0.58} \\
\bottomrule
\end{tabular}
\end{sc}
\end{scriptsize}
\end{center}
\vskip -0.2in
\end{table}
\begin{table}[hbt!]
\caption{
The effect of the completeness requirement (Algorithm 1, line 18) on the structural correctness. The numbers of extra and missing edges in the graph resulting from HCCD, are divided by the corresponding values after relaxing the completeness requirement. Values represent ratios, e.g., a value of 0.1 indicates that only 10\% of errors in the non-complete algorithm are present in HCCD.
}
\vskip -0.15in
\label{tab:acc_completeness}
\begin{center}
\begin{scriptsize}
\begin{sc}
\begin{tabular}{lccccc}
\toprule
Density & $\rho=3$ & $\rho=4$ & $\rho=5$ & $\rho=6$ & $\rho=7$\\
\midrule

Extra edges ratio & 0.11 &  0.08  & 0.10    & 0.16 & 0.21 \\
Missing edges ratio & 1.11  & 1.12    & 1.12    & 1.10 & 1.08\\
\bottomrule
\end{tabular}
\end{sc}
\end{scriptsize}
\end{center}
\vskip -0.4in 
\end{table}

\section{An Analysis using Synthetic Data}
\begin{figure}[h!]
    \centering
    \subfigure[]{\includegraphics[width=0.21\textwidth]{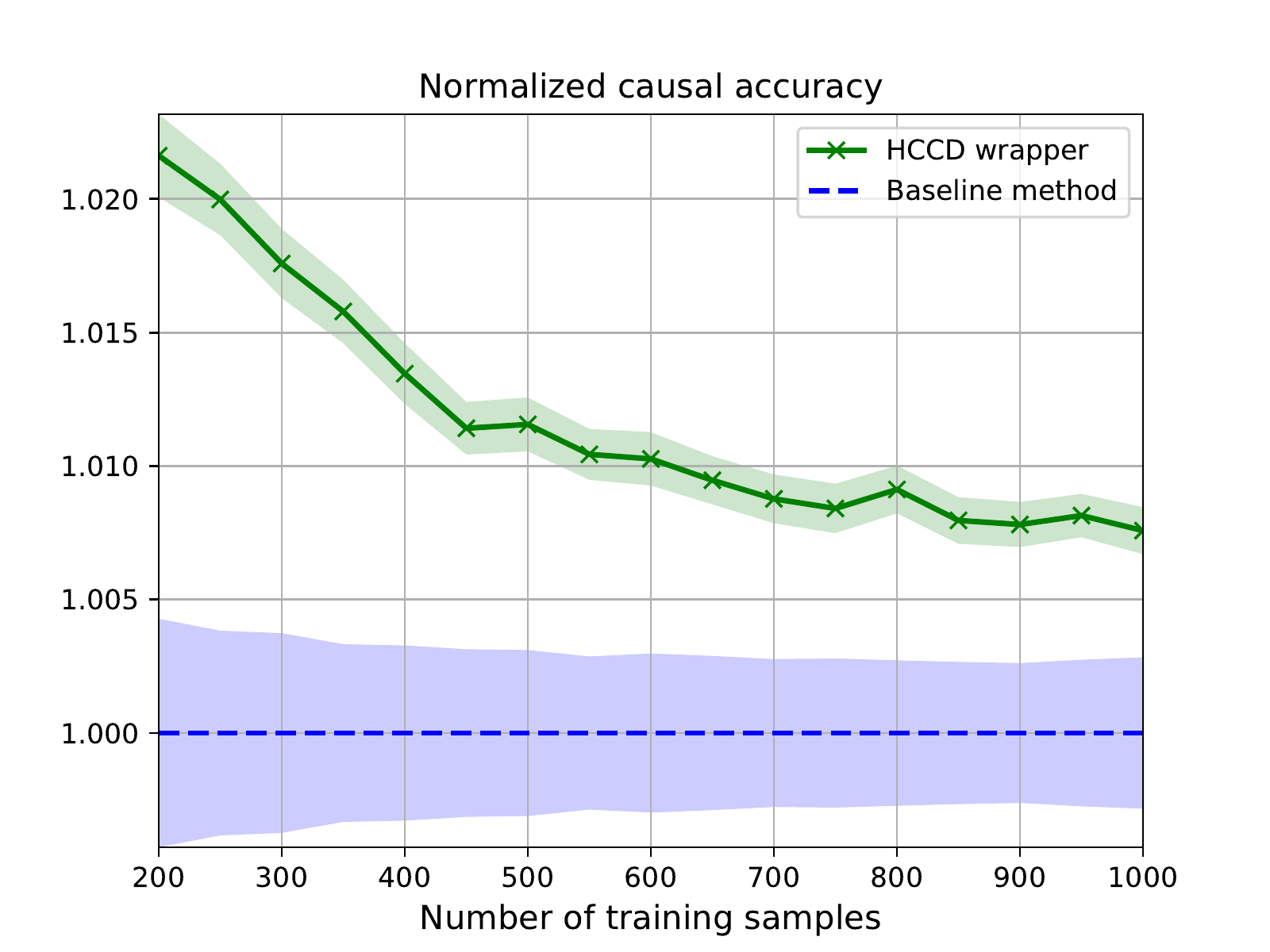}} 
    \subfigure[]{\includegraphics[width=0.21\textwidth]{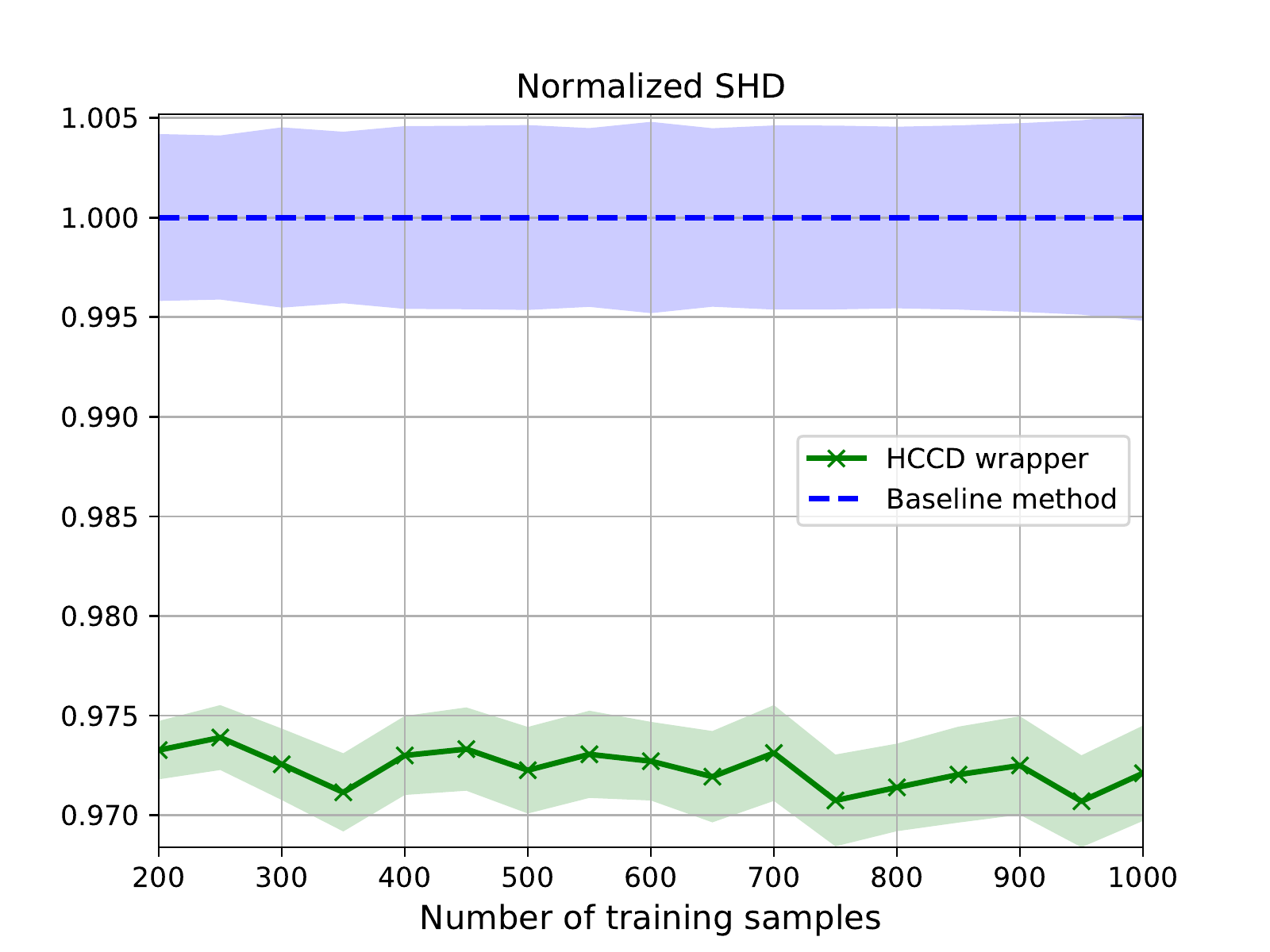}}
    \vskip -0.05in
    \subfigure[]{\includegraphics[width=0.21\textwidth]{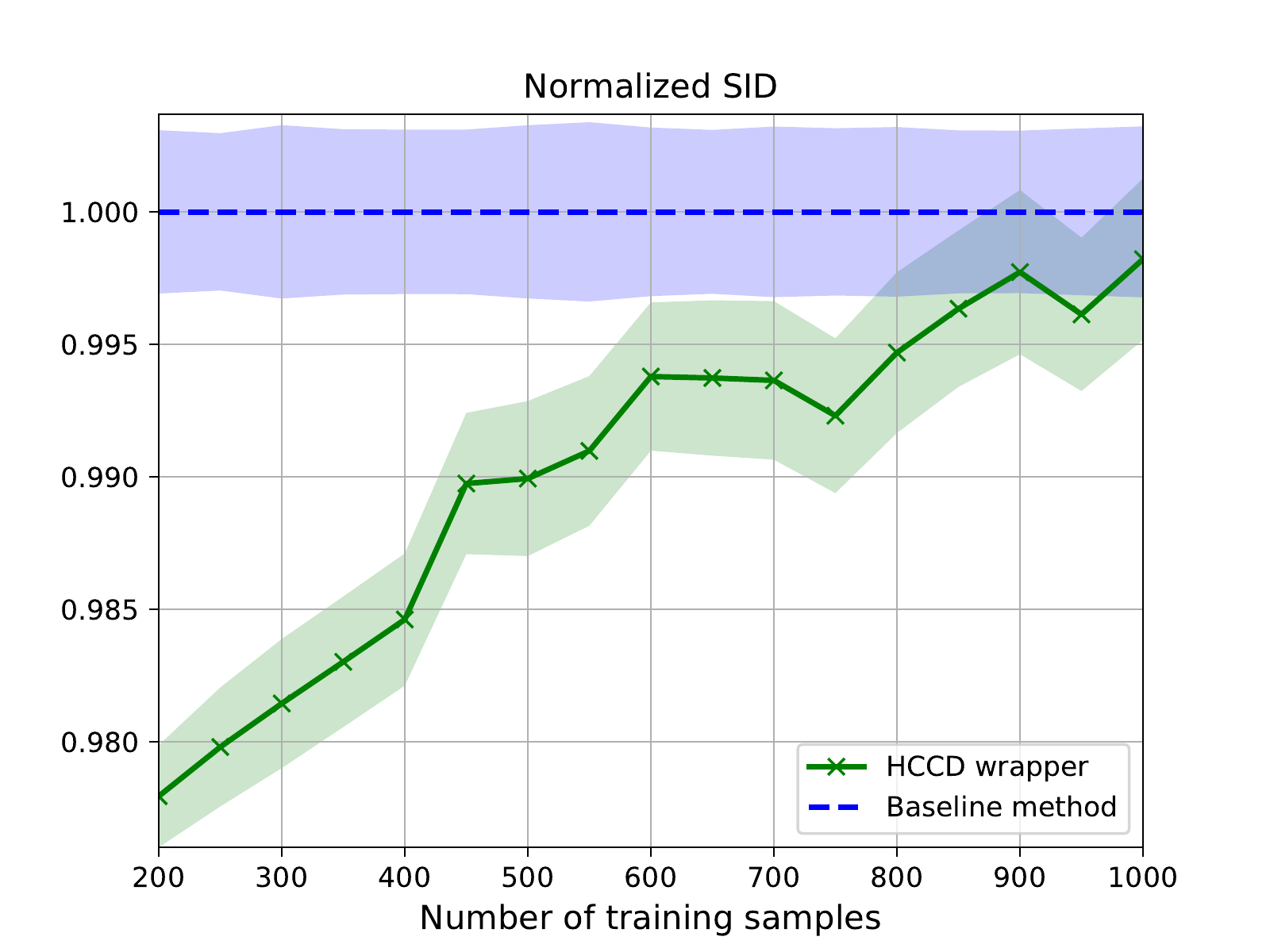}}
    \subfigure[]{\includegraphics[width=0.21\textwidth]{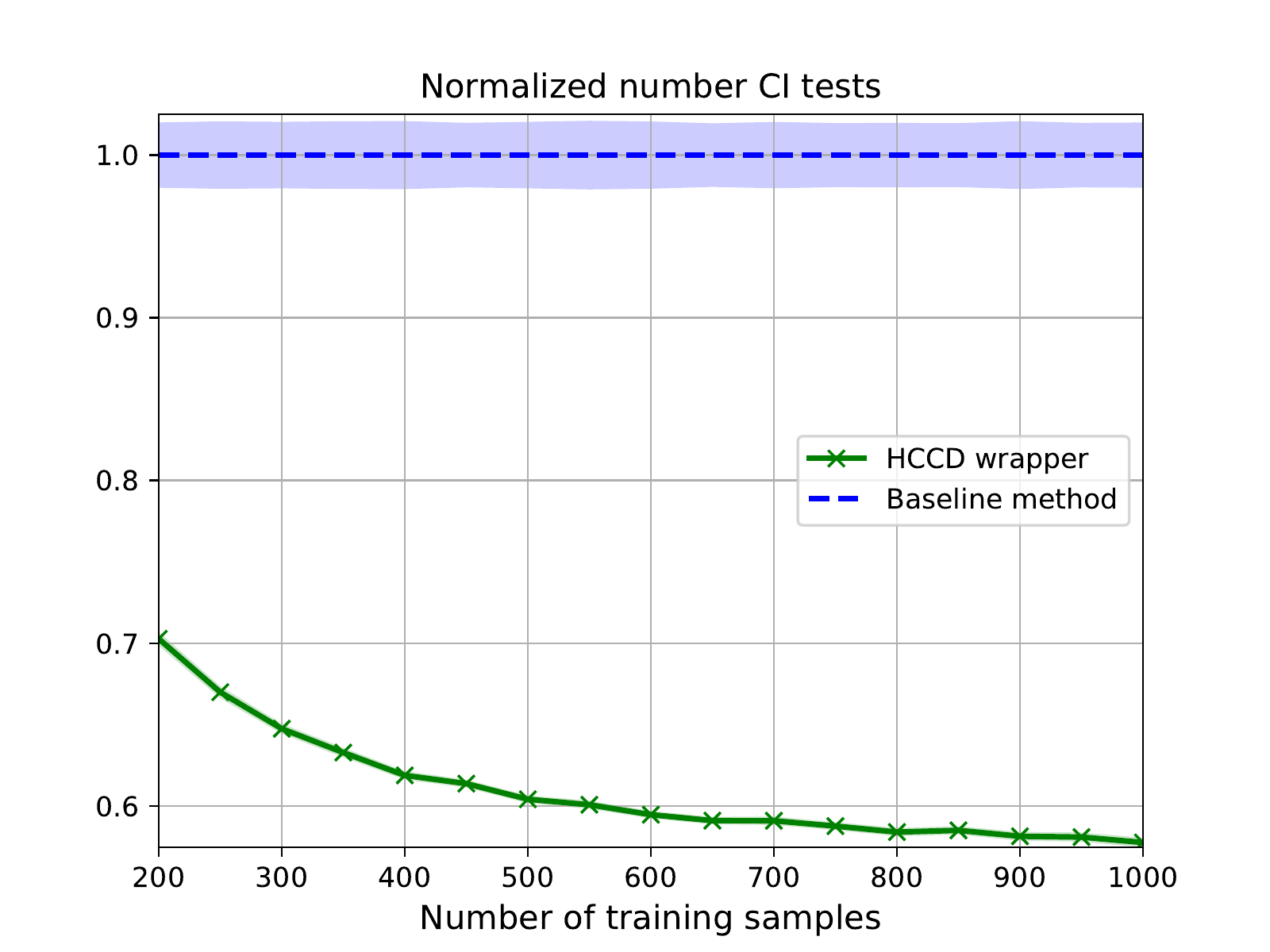}}
    \vskip -0.05in
    \subfigure[]{\includegraphics[width=0.21\textwidth]{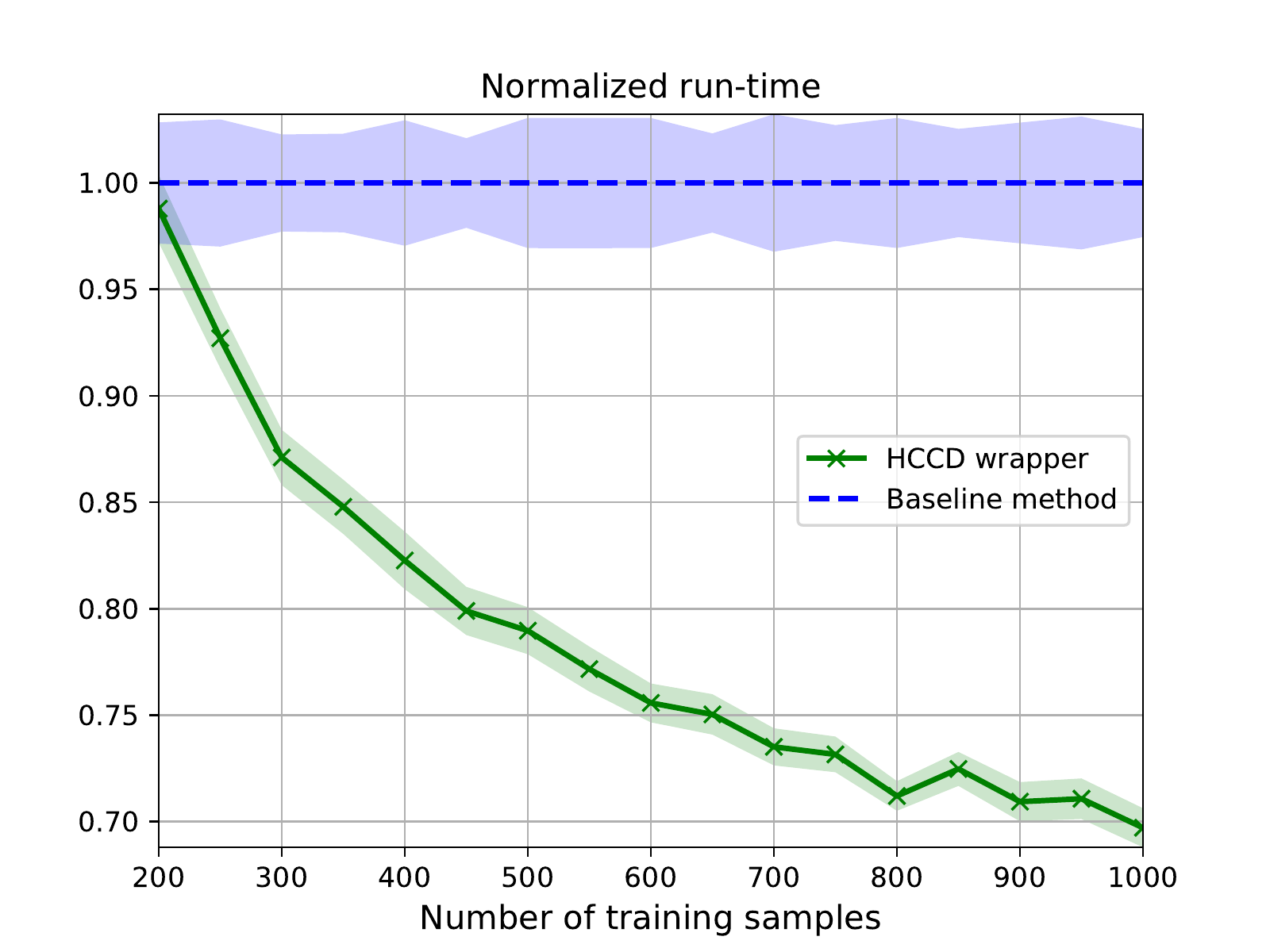}}
    \vskip -0.15in

    \caption{Performance of the HCCD wrapper, relatively to the baseline (PC), as a function of the number of training samples, for 200 graph nodes.}
    \vskip 0.12in
    \label{fig:scalability_to_training_set_200}
\end{figure}

\begin{figure} [ht!]
    \centering
    \subfigure[]{\includegraphics[width=0.21\textwidth]{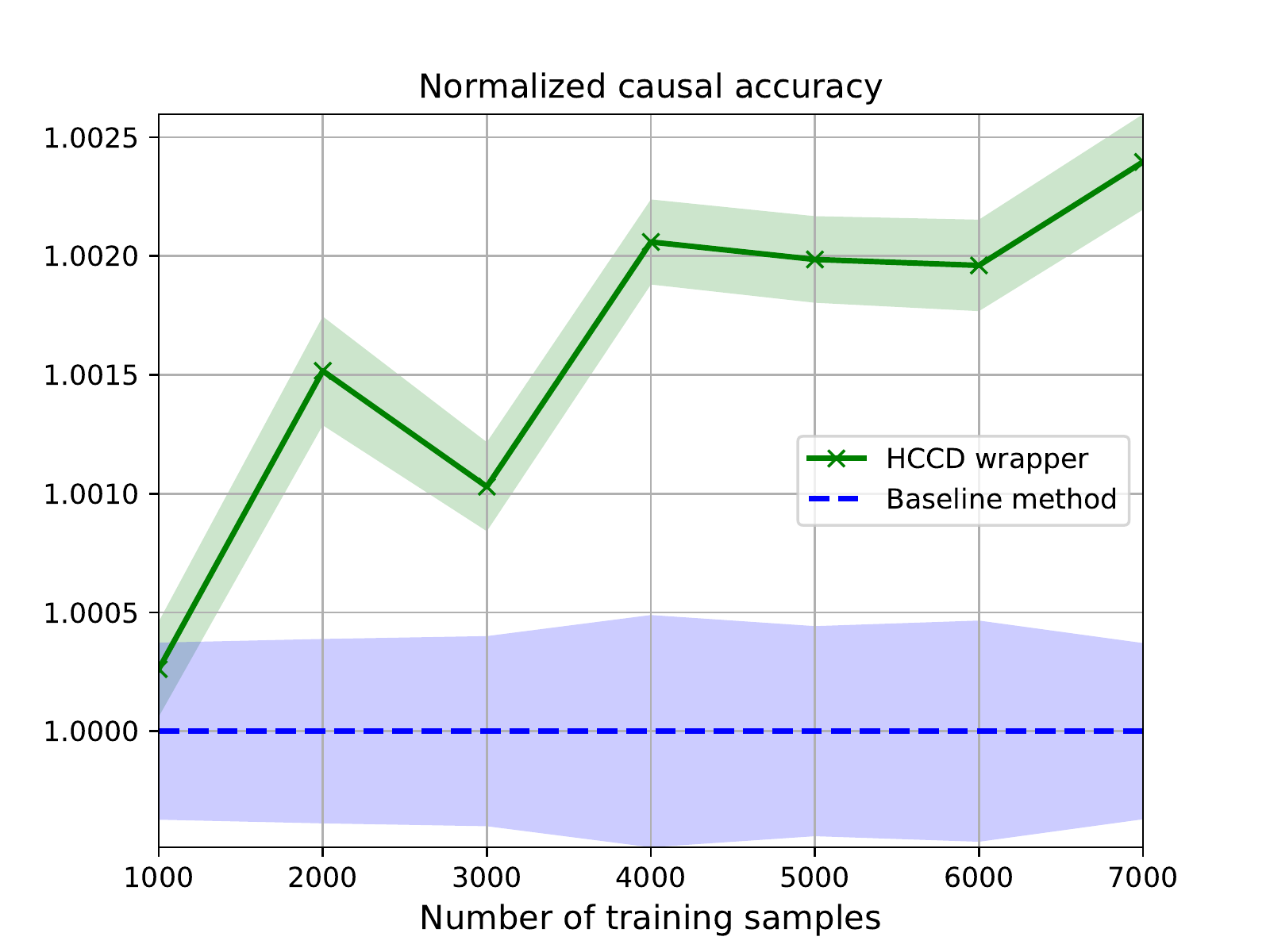}} 
    \subfigure[]{\includegraphics[width=0.21\textwidth]{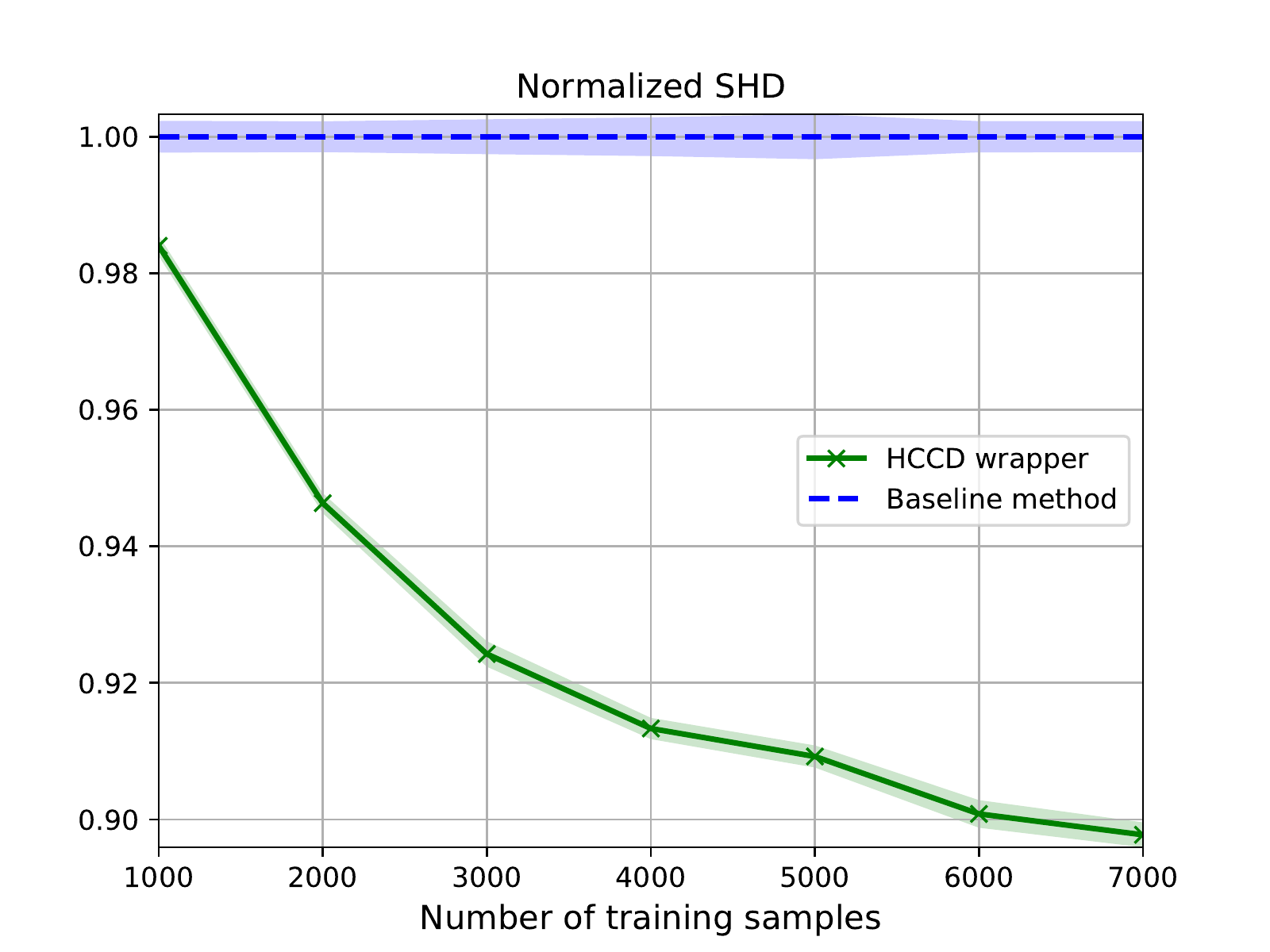}}
    \vskip -0.05in
    \subfigure[]{\includegraphics[width=0.21\textwidth]{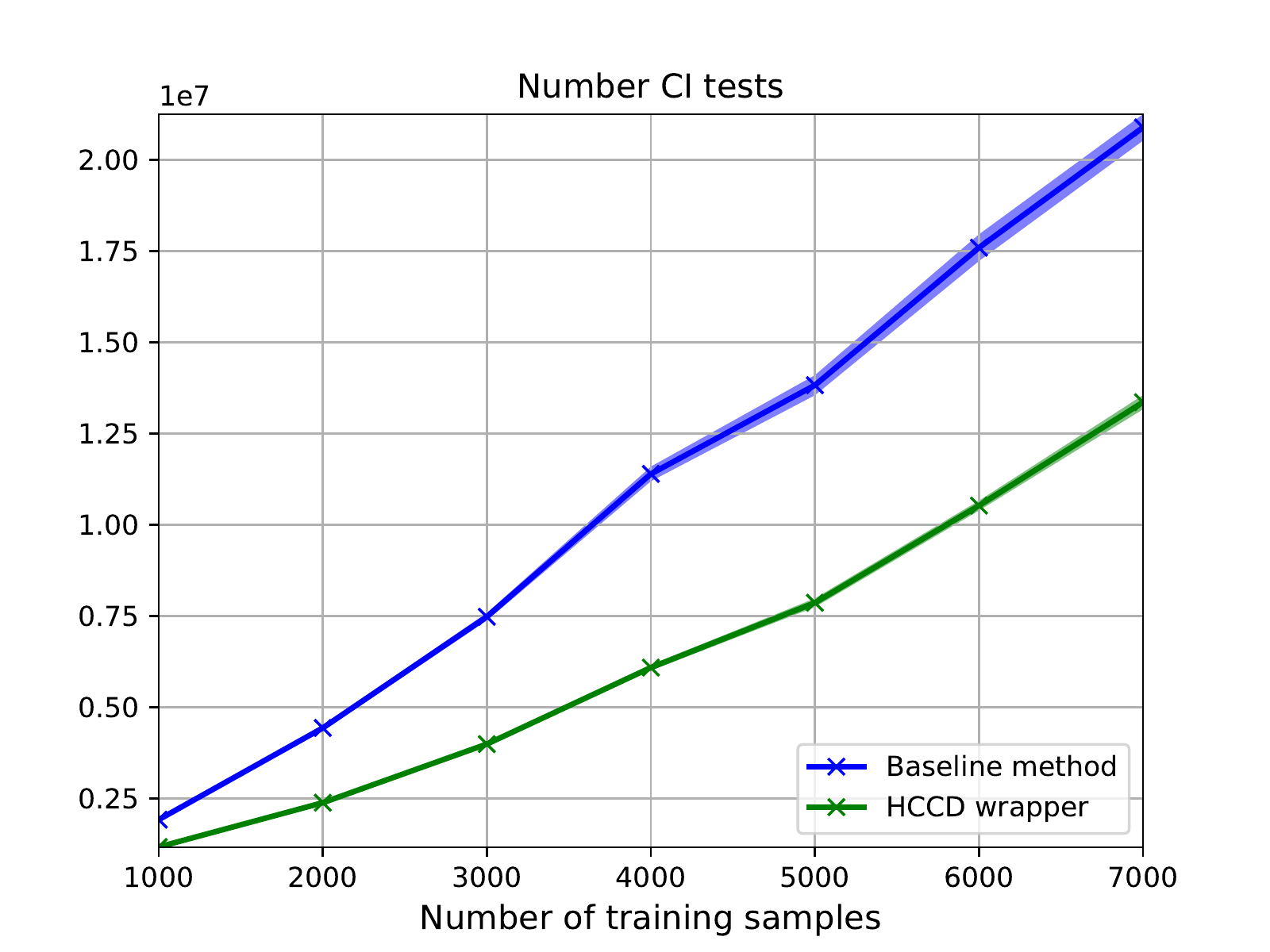}}
    \subfigure[]{\includegraphics[width=0.21\textwidth]{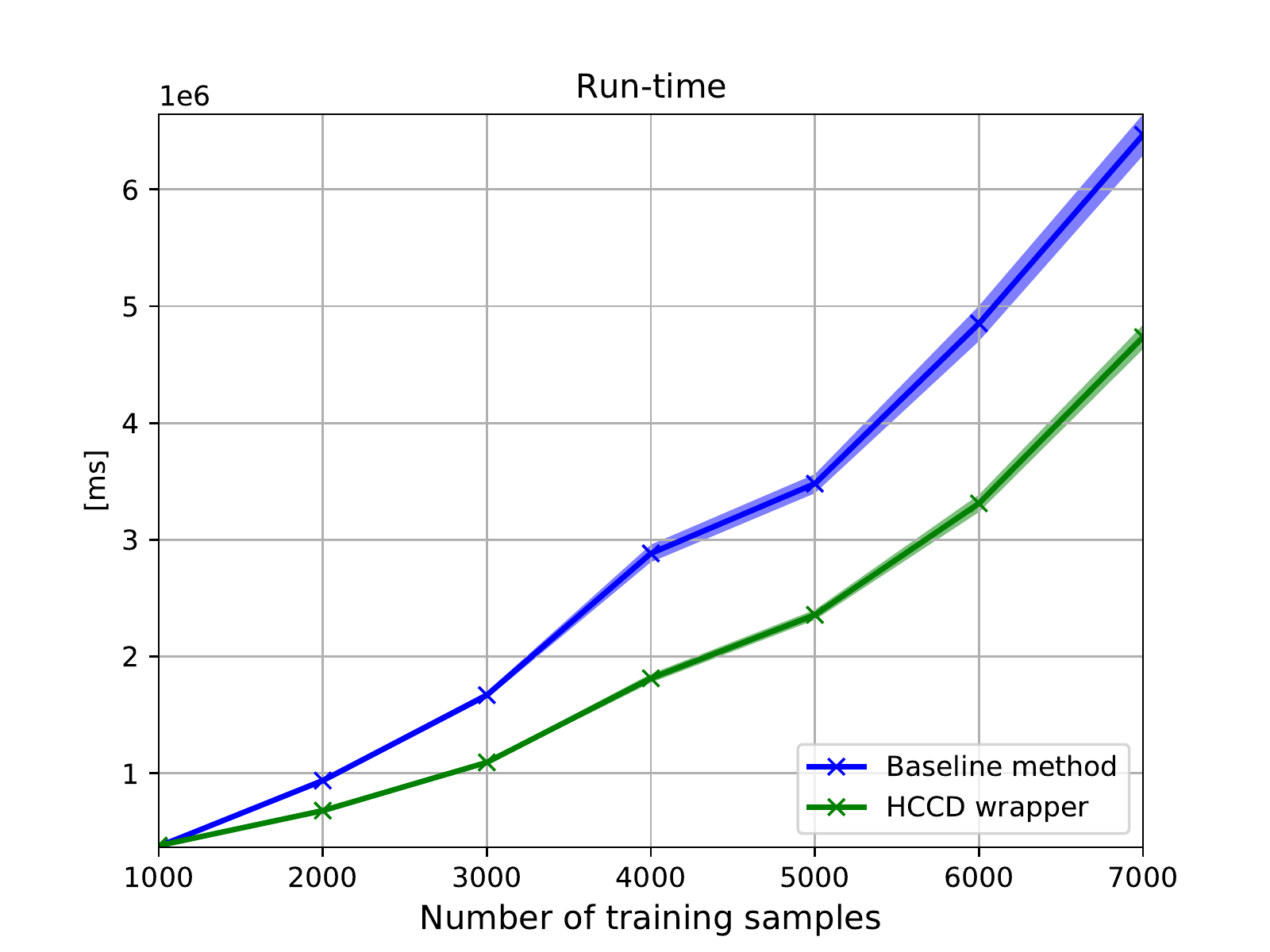}}

    \vskip -0.15in

    \caption{Performance of the HCCD wrapper, relatively to the baseline (PC), as a function of the number of training samples, for 1000 graph nodes. In (c) and (d) we present the absolute values in order to demonstrate the actual savings.}
    \label{fig:scalability_to_training_set_1000}
\end{figure}

\vskip -0.1in
Section 4.2 evaluates the performance of the HCCD wrapper with respect to the number of training set size, and to the number of nodes in the graphs, compared to the baseline method (PC) over 5 measures, for graphs with $n = 100$ nodes. This evaluation is extended here, in  \Crefrange{fig:scalability_to_training_set_200}{fig:scalability_to_training_set_1000}, to other (low and high) number of nodes in a graph, i.e. $n\in\{200,1000\}$. Each figure shows mean $\pm$ std of 500 independent tests (DAGs), and values are normalized by the PC score in order to analyze the improvement over the baseline method. In each figure, the evaluated metrics are: (a) Causal accuracy (higher is better); (b) SHD (lower is better); (c) SID (lower is better); (d) Number of CI tests (lower is better). (e) Run-time (lower is better). It is evident that the HCCD performs well for all 3 structural correctness metrics along the entire range of the training set size and for every $n$, and it is superior to the baseline method. In addition, we can see the saving in the number of CI tests, and importantly the reduction in the run-time (which includes the clustering stage) for the entire range of the training set size. One exception is for the case of $n=20$ (figure not presented here due to lack of space), for which the HCCD run time is higher, which is expected since the complexity overhead of the clustering stage overtakes the saving in run-time gained by using less statistical tests in datasets with a small number of nodes. Nevertheless, for the common real-world cases of datasets with many domain variables, the HCCD achieves a significant reduction in run-time. Additionally, it is evident that the run-time reduction increases with the increase of the number of training samples, i.e. bigger training sets gain more in real-time reduction.

\vskip -0.1in

\vskip -0.15in

\end{document}